\documentclass[10pt]{article}

\usepackage[margin=1in]{geometry}
\usepackage[utf8]{inputenc} 
\usepackage[T1]{fontenc}    
\usepackage{url}            
\usepackage{booktabs}       
\usepackage{amsfonts}       
\usepackage{nicefrac}       
\usepackage{microtype}      
\usepackage{xcolor}         
\usepackage{graphicx}
\usepackage{amsmath}
\usepackage{amssymb}
\usepackage{caption}
\usepackage{subfigure}
\usepackage{amsmath}
\usepackage{amssymb}
\usepackage{float}
\usepackage{algorithmic}
\usepackage{algorithm}
\usepackage[round,semicolon]{natbib}

\usepackage{url}            
\usepackage{booktabs}       
\usepackage{amsfonts}       
\usepackage{nicefrac}       
\usepackage{microtype}      
\usepackage{longtable}
\usepackage{tgpagella}
\usepackage{listings}
\usepackage{color}

\definecolor{dkgreen}{rgb}{0,0.6,0}
\definecolor{gray}{rgb}{0.5,0.5,0.5}
\definecolor{mauve}{rgb}{0.58,0,0.82}

\usepackage{multicol}

\usepackage{hyperref}

\definecolor{darkblue}{rgb}{0, 0.2, 0.7}
\hypersetup{
    colorlinks = true,
    linkcolor = darkblue,
    anchorcolor = darkblue,
    citecolor = darkblue,
    filecolor = darkblue,
    urlcolor = darkblue
}

\usepackage{cleveref}

\begin{document}

\title{\vspace{-2em}%
  \hrule height 4pt%
  \vskip 0.25in%
  \vskip -\parskip%
  \textbf{UL2: Unifying Language Learning Paradigms
}
  \vskip 0.2in%
  \vskip -\parskip%
  \hrule height 1pt%
  \vskip 0.09in}

\author{%
 \textbf{Yi Tay}\thanks{Yi and Mostafa are co-leads of this project and are denoted with $^*$. $\sharp$ denotes technical research contributors. $\flat$ denotes data \& infrastructure contributions. $^\triangle$ denotes advising contributions. Don, denoted with $^\square$ is the last author. Full contributions of all authors at the end of paper. Correspondence to \url{yitay@google.com} or \url{dehghani@google.com}.} \hspace{5mm} \textbf{Mostafa Dehghani}$^*$  \vspace{5mm} \\ \hspace{3mm}  \textbf{Vinh Q. Tran}$^\sharp$  \hspace{3mm} \textbf{Xavier Garcia}$^\sharp$ \hspace{3mm}  \textbf{Jason Wei}$^\sharp$ \hspace{2mm} \textbf{Xuezhi Wang}$^\sharp$ \hspace{2mm}  \textbf{Hyung Won Chung}$^\sharp$ \hspace{3mm} \vspace{5mm}   \\ 
 \textbf{Siamak Shakeri}$^\sharp$ \hspace{3mm} \textbf{Dara Bahri}$^\flat$  \hspace{3mm}  \textbf{Tal Schuster}$^\flat$ \hspace{3mm}  \hspace{3mm}  \textbf{Huaixiu Steven Zheng}$^\triangle$ \hspace{3mm} \vspace{5mm} \\
  \textbf{Denny Zhou}$^\triangle$ \hspace{3mm} \textbf{Neil Houlsby}$^\triangle$ \hspace{3mm} \textbf{Donald Metzler}$^\triangle$ \vspace{8mm} \\
Google Brain \vspace{6mm} \\ 
}

 

\date{}

\maketitle
\begin{abstract}
Existing pre-trained models are generally geared towards a particular class of problems. To date, there seems to be still no consensus on what the right architecture and pre-training setup should be. This paper presents a unified framework for pre-training models that are universally effective across datasets and setups. 
We begin by disentangling architectural archetypes with pre-training objectives -- two concepts that are commonly conflated. Next, we present a generalized and unified perspective for self-supervision in NLP and show how different pre-training objectives can be cast as one another and how interpolating between different objectives can be effective. We then propose Mixture-of-Denoisers (MoD), a pre-training objective that combines diverse pre-training paradigms together. We furthermore introduce a notion of mode switching, wherein downstream fine-tuning is associated with specific pre-training schemes. We conduct extensive ablative experiments to compare multiple pre-training objectives and find that our method pushes the Pareto-frontier by outperforming T5 and/or GPT-like models across multiple diverse setups. Finally, by scaling our model up to 20B parameters, we achieve SOTA performance on 50 well-established supervised NLP tasks ranging from language generation (with automated and human evaluation), language understanding, text classification, question answering, commonsense reasoning, long text reasoning, structured knowledge grounding and information retrieval. Our model also achieve strong results at in-context learning, outperforming 175B GPT-3 (published paper results) on zero-shot SuperGLUE and tripling the performance of T5-XXL on one-shot summarization. On zero-shot MMLU, UL2 20B outperforms T0 and T5 models. Additionally, we show that UL2 20B works well with chain-of-thought prompting and reasoning, making it an appealing choice for research into reasoning at a small to medium scale of 20B parameters. Finally, we apply FLAN instruction tuning to the UL2 20B model, achieving MMLU and Big-Bench scores competitive to FLAN-PaLM 62B. We release Flax-based T5X model checkpoints for the UL2 20B model and Flan-UL2 20B model at \url{https://github.com/google-research/google-research/tree/master/ul2}.

\end{abstract}
\newpage

\tableofcontents
\newpage
\section{Introduction}

There is a wide spectrum of pre-trained model options for NLP researchers and practitioners these days \citep{devlin2018bert,brown2020language,raffel2019exploring,Radford2019,liu2019roberta,yang2019xlnet, thoppilan2022lamda,fedus2021switch,du2021glam,chowdhery2022palm}. When faced with the question of what model should one use, the answer is often \textit{it depends}, followed by \textit{on what task?}

Answering this can be overwhelming, comprising of a number of fine-grained follow-up questions like, \textit{`encoder-only or encoder-decoder?'}, \textit{`span corruption or language model?'}. Pressing further, the answer always seems to depend on the target downstream task. This paper questions and rethinks this thought process, specifically answering the questions of \textit{why should the choice of the pre-trained LM depend on the downstream task?} and \textit{how can we pre-train models that work universally well across many tasks?}.

\begin{figure}[H]
    \centering
    \includegraphics[width=0.65\linewidth]{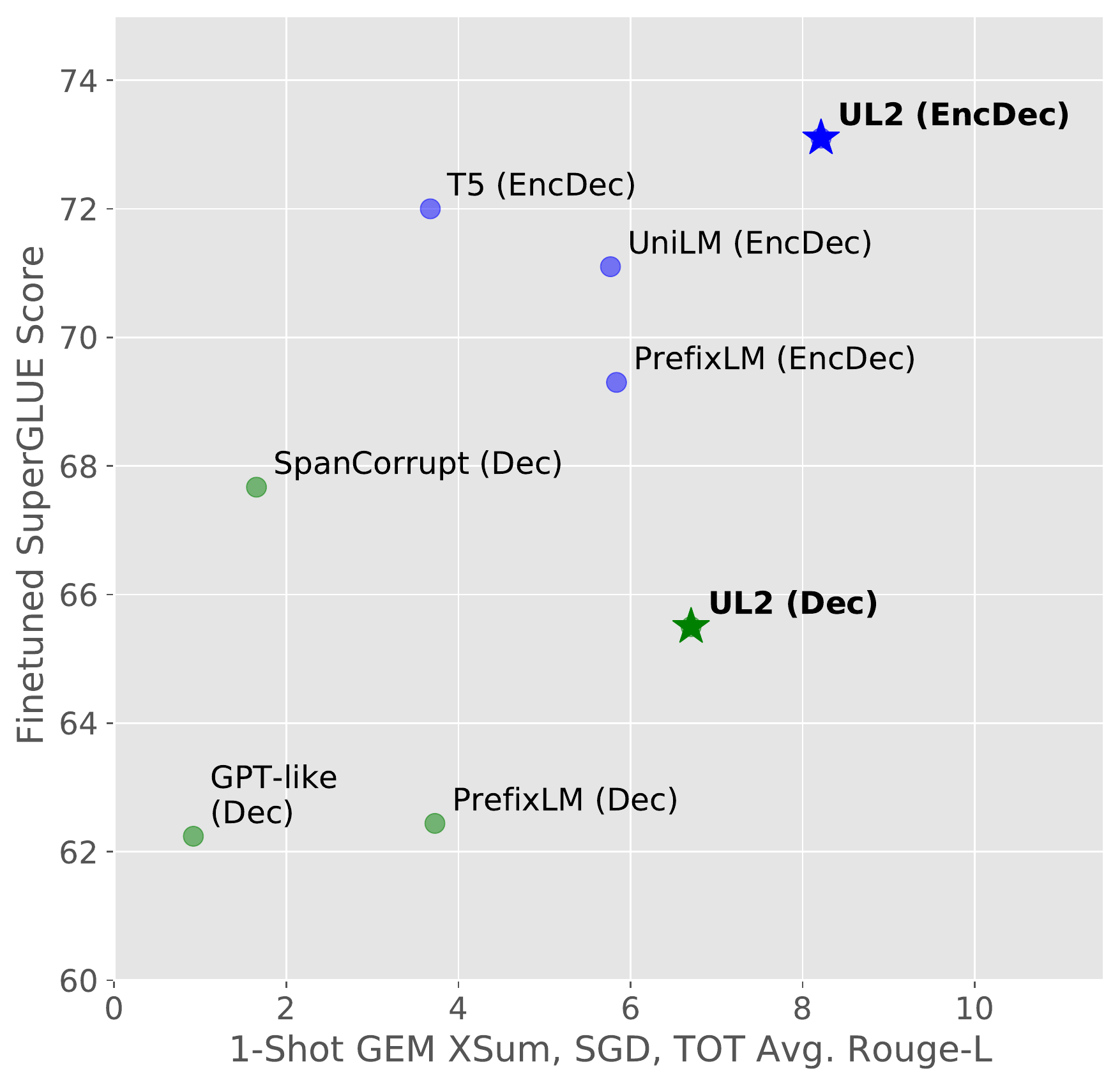}
    \captionsetup{width=\linewidth}
\caption{In both decoder-only and encoder-decoder setups, UL2 strikes a significantly improved balance in performance between fine-tuned discriminative tasks and prompt-based 1-shot open-ended text generation than previous methods. Note: Dec and EncDec are compute matched but EncDec models have double the parameters.}
  \label{fig:UL2-scatter}
\vspace{-1em}
\end{figure}

This paper proposes a step towards making a universally applicable language model possible. We present a framework for \textit{Unifying Language Learning Paradigms} or UL2 in short, that is consistently effective across a very diverse set of tasks and setups. Figure \ref{fig:UL2-scatter} shows an example of how UL2 can perform universally well, unlike other models that often have to make a trade-off.

The appeal of a universal model is clear, i.e., as this not only allows concentrated effort in improving and scaling a single model, instead of diversifying resources across $N$ models. Moreover, under resource constrained settings where only a few models can be served (e.g., on device), it would be preferable to have a single pretrained model that can perform well on many types of tasks.

At the core of UL2 is a the newly proposed Mixture-of-Denoisers (MoD), a pre-training objective that enables strong performance across tasks. MoD is a mixture of several well-established denoising objectives along with new ones; namely X-denoising (extreme denoising) which considers extreme span lengths and corruption rates, S-denoising (sequential denoising) that strictly follows sequence order, and R-denoising (regular denoising) that is a standard span corruption objective introduced in \citep{raffel2019exploring}. We show that MoD is conceptually simple but highly effective for a diverse set of tasks.

Our approach exploits the realization that most (if not all) well-studied pre-training objectives differ in the type of context a model is conditioned on. For example, the span corruption objective is akin to invoking multiple regions of prefix language modeling (PLM)  \citep{liu2018generating,raffel2019exploring} whereby prefixes are contiguous segments of non-corrupted tokens and targets have full access to prefixes of all PLM segments. The setting where the span approaches the full sequence length is approximately a language modeling objective conditioned on long-range context. Thus, we are able to design a pre-training objective that smoothly interpolates these different paradigms (span corruption vs language modeling vs prefix language modeling). 

It is also easy to see that each denoiser is difficult in different ways. They also differ in the nature of \textit{extrapolation} (or interpolation). For example, bounding a model by bidirectional context (or the future) (ie.., span corruption) makes the task easier and becomes more akin to fact completion. Meanwhile, PrefixLM/LM objectives are generally more \textit{`open ended`}. These behaviours can be easily observed by monitoring the cross entropy losses of these different denoising objectives. 

Given the MoD formulation, we conjecture that it is beneficial for our model to not only distinguish between different denoisers during pre-training but also to adaptively switch modes when learning downstream tasks. We introduce mode switching, a new concept that associates pre-training tasks with dedicated sentinel tokens and allows dynamic mode switching via discrete prompting. Our model is able to switch modes between R,S and X denoisers on-demand after being pre-trained.

We then disentangle the architecture from the self-supervision scheme. While it might be a common misconception, as previously noted in \citet{raffel2019exploring}, that a pre-trained model is strongly characterized by its backbone architecture (e.g., decoder-only vs.\ encoder-decoder), we find that the choice of the denoiser has significantly more impact. MoD supports either backbone, similar to how T5's span corruption may be trained with a decoder-only model. As such, UL2 is agnostic to architecture. We argue that the choice of backbone architecture is mainly a trade-off across different efficiency metrics.

We conduct systematic and ablative experiments on a suite of 9 diverse tasks aimed to capture different problem formulations (supervised and prompt-based in-context few-shot learning). We experiment with the SuperGLUE suite \citep{wang2019superglue}, and three tasks from the GEM benchmark \citep{gehrmann2021gem}. In addition, we evaluate on open text generation, as well as prompt-based one-shot settings on all tasks. In this ablative setup, our experimental results show that UL2 outperforms T5 and GPT-like baselines on all 9 setups. On average, UL2 outperforms a T5 baseline by $+43.6\%$ and a language model by $+76.1\%$. Among all the other competitive baselines considered, UL2 is the only method that outperforms T5 and GPT-like models on all tasks. 

We scale UL2 up to a moderate scale setting of approximately 20B (19.5 to be exact) parameters and run experiments across a very diverse suite of 50+ NLP tasks ranging from language generation (with automated and human evaluation), language understanding, text classification, question answering, commonsense reasoning, long text reasoning, structured knowledge grounding and information retrieval. Our results show that UL2 achieves SOTA on a vast majority of tasks and setups. 

Finally, we conduct zero/few-shot experiments with UL2 and show that UL2 outperforms GPT-3 175B on zero shot SuperGLUE. When compared with newer state-of-the-art models like GLaM \citep{du2021glam}, PaLM \citep{chowdhery2022palm} and ST-MoE \citep{zoph2022designing}, UL2 remains competitive at a compute-matched setup despite only training on C4 corpus which is known to be less effective than specially curated datasets used in \citep{du2021glam,chowdhery2022palm}. We delve into understanding trade-offs between zero-shot and finetuning performance and show that UL2 is Pareto-efficient with respect to both learning paradigms. On one-shot summarization, UL2 triples the performance of an LM adapted T5 XXL model and is competitive with (or outperforms) PaLM and LaMDA at the same compute cost. We release T5X-based Flax checkpoints of the trained UL2 model.  
\newpage

\section{Background: Pre-trained Language Models}
In this section, we discuss background surrounding pretrained language models, pretraining objectives and other unified pretraining proposals.

\subsection{Pre-trained Language Models}
Learning pre-trained representations for language is a far-reaching pillar of modern NLP research, dating back to \citep{mikolov2013distributed,pennington2014glove,neumann2018deep,dai2015semi,howard2018universal}. The first pre-trained Transformer, GPT, was proposed by \citep{Radford2019} and was trained as a causal language model. Subsequently, BERT \citep{devlin2018bert} demonstrated the importance of bidirectional modeling for many downstream tasks. BERT introduced masked language modeling (MLM), a denoising objective that reconstructs the input in-place using bidirectional receptive fields. XLNet \cite{yang2019xlnet} introduced the Permutation Language Modeling to account for dependencies between masked tokens during training. A number of additional papers (e.g., RoBERTA~\citep{liu2019roberta}, SpanBERT~\citep{joshi2020spanbert}) suggested further improvements to the pre-training process. 

At the same time, two-stack encoder-decoder architectures such as T5~\citep{raffel2019exploring} gained popularity due to their improved performance on classification and sequence-to-sequence (``seq2seq'') tasks. However, so far, these models have shown limited performance on open-text generation and prompt-based inference (i.e., in-context learning), which motivates the use of decoder-only models that are trained with different objectives (e.g., GPT-3~\citep{brown2020language}, GLaM~\citep{du2021glam}, LaMDa~\citep{thoppilan2022lamda} and PaLM \citep{chowdhery2022palm}). In this work, we aim to bridge the performance gap between the two by means of a general training paradigm that suits both architectures.

\paragraph{Decoder-only vs Encoder-only} The key similarities of decoder-only and encoder-only architectures is that decoder-only architectures operate with an \textit{input-to-target} paradigm or \textit{targets-only} paradigm if CausalLM is used over PrefixLM used. For both architectures, the objective is always to predict the next token (LM) and are therefore autoregressive models. Notably this is different from position-wise masked LM denoising (sometimes known as \textit{autoencoding}), which have been popularized by encoder-only BERT-style models. These class of models are very restricted in their generative capabilities. On top of that, task specific classification heads are also typically employed for downstream tasks. Because of the cumbersomeness of task specific classification heads, we strongly do not recommend using this class of autoencoding models moving forward and consider them somewhat deprecated. Caveats do apply. For instance, regression is the probably only reason why one would add a task specific head \citep{lees2022new}, or to squeeze out some efficiency gains from eliminating a full vocabulary. Either way, one can always start from a encoder-decoder and chop off the decoder later so there is no good reason to use an encoder-only model. Hence the only real objective consideration here is between decoder-only and encoder-decoder architectures.

\paragraph{Decoder-only vs Encoder-Decoder}
The line between decoder-only and encoder-decoder models is less clear. PrefixLM models are \textit{almost} encoder-decoder models with shared parameters (but not quite). From an inductive bias point of view, there are multiple differences. Encoder-Decoder models process input and targets independently with a different set of parameters. This is a form of sparsity where different set of parameters are used for different tokens. Encoder-Decoder models also have a cross attention component that connects input tokens to target tokens. Meanwhile, decoder-only models process inputs and targets by concatenating them. Hence, the representations of inputs and targets are concurrently build layer by layer as the input/targets propagate up the network. Conversely, the decoder in Encoder-decoder models generally only looks at the fully processed encoder input. Overall, the inductive bias of PrefixLM decoder-only models and Encoder-Decoder models could be pretty similar modulo the subtle differences stated above. The distinct property is that Encoder-Decoder models are generally approximately 2x parameters of a decoder-only model when compute-matched.

\paragraph{Sparse Models} On a side note, there have also been also an emerging trend of sparse pretrained models that achieve state-of-the-art performance. Sparse mixture-of-expert models such as the Switch Transformer \citep{fedus2021switch}, GLaM \citep{du2021glam} and/or GShard \citep{lepikhin2020gshard} have also demonstrated a lot of promise. While orthogonal to the topic of pretraining objectives, sparse models achieve a very different flop-per-parameter ratio compared to dense models - a core recurring motif in the debate surrounding encoder-decoder models vs decoder-only models.

\subsection{Pre-training Objectives for Large Language Models}
While recent research demonstrates the potential of large \emph{supervised} multi-task pre-training \citep{aribandi2021ext5,sanh2021multitask,wang2022language}, most pre-training objectives rely on the vast availability of \emph{unsupervised} data and use self-training techniques.
As mentioned above, different architectures typically leverage different objectives. Decoder-only models are typically trained with causal language model objectives to mimic auto-regressive generation \citep{Radford2019}. \citet{raffel2019exploring} explored many objectives for encoder-decoder models and found span corruption to be effective. \citep{wang2022language} conducts a systematic study of different architectures combined with three different pretraining objectives (causal LM, prefixLM and span corruption) and analyzed their impact on zero-shot generalization. Related to our proposed X-denoisers, \citep{wettig2022should} studies the effect of corruption rate in BERT-style masked language modeling and hypothesizes that this improves sample efficiency along with benefitting larger models. Notably, the benefits of heightened corruption rates as a \textit{standalone} denoiser is still unclear, as noted by \citep{raffel2019exploring} and also apparent in our own ablations. Pre-training (or denoising) is generally applied on the subword level \citep{raffel2019exploring,devlin2018bert} but it is worth to note that it has also been applied on the character or byte-level \citep{xue2021byt5,tay2021charformer}. In these setups, the corrupted spans are generally much larger than subword-based denoising.


\subsection{Unified Pre-training Proposals}
UniLM \citep{dong2019unified} proposed to train on multiple language modeling objectives using a single Transformer model. Specifically, UniLM trains on unidirectional LM, bidirectional LM and seq2seq LM. This is quite similar to combining auto-regressive LMs with BERT and prefix-LM models. Notably, UniLM trains using a cloze-type formulation which adds explicit mask tokens to the inputs. Losses are then computed by the difference of the predicted token and target token in a position-wise fashion. Aside from pretraining unification, there have been a recent trend of thematic unification, i.e., unifying common tasks into one model. Examples of these include UNICORN \citep{lourie2021unicorn} for commonsense reasoning, UnifiedQA \citep{khashabi2020unifiedqa,khashabi2022unifiedqa} for question answering and UnifiedSKG \citep{xie2022unifiedskg} for Structured Knowledge Grounding.




\section{Unifying Language Learning Paradigms (UL2)}
This section describes the UL2 framework and the proposed pre-training objectives which we study for the remainder of the paper.

\subsection{Pre-training}
This section discusses the proposed pre-training objective.


\begin{figure*}[t!]
    \centering
    \includegraphics[width=1.0\linewidth]{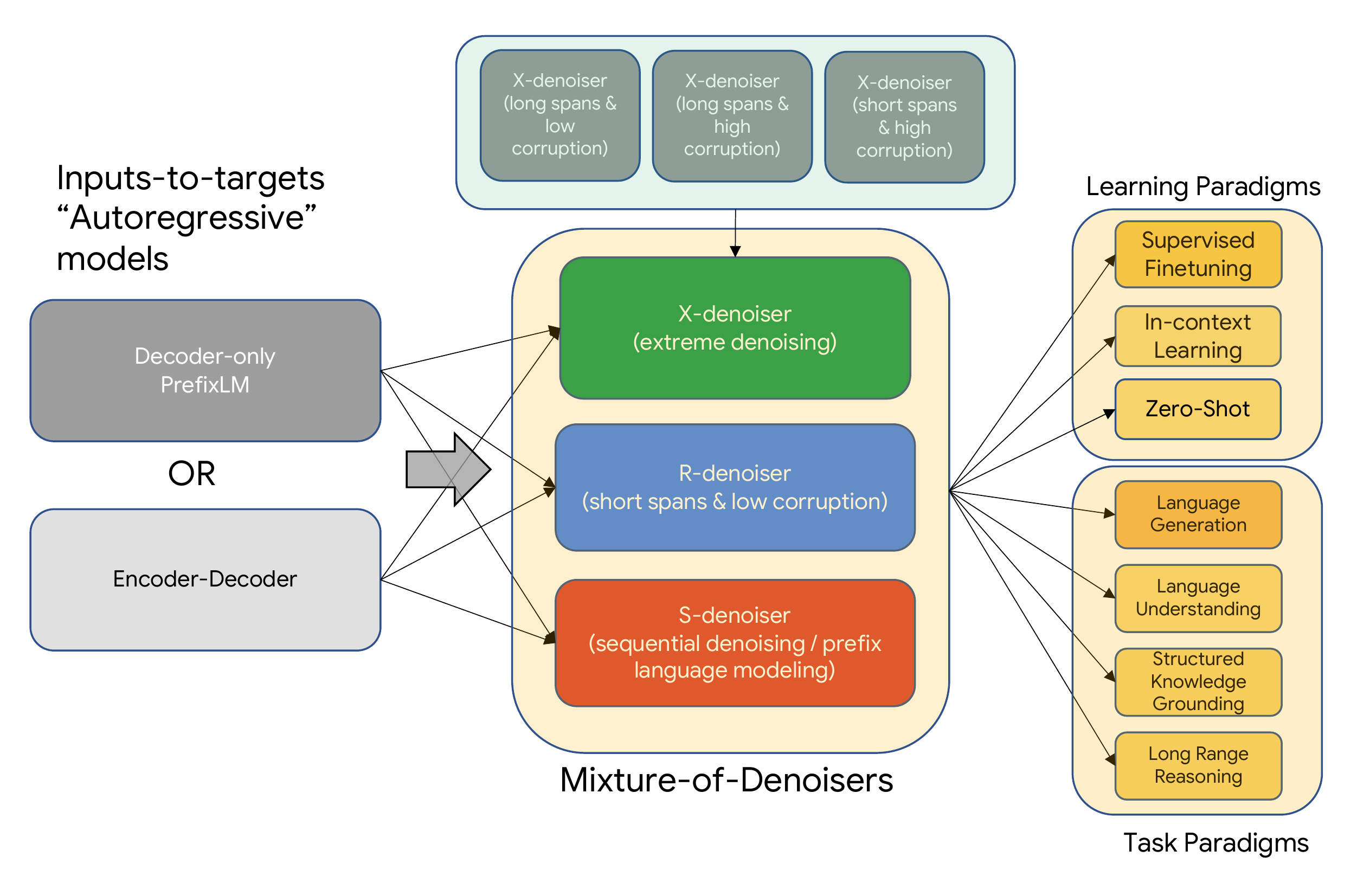}
        \vspace{-1em}
    \label{fig:ul2_overview}
\caption{An overview of UL2 pretraining paradigm. UL2 proposes a new pretraining objective that works well on a diverse suite of downstream tasks.}
\label{stats}
\end{figure*}
\begin{figure*}[t!]
    \centering
    \includegraphics[width=1.0\linewidth]{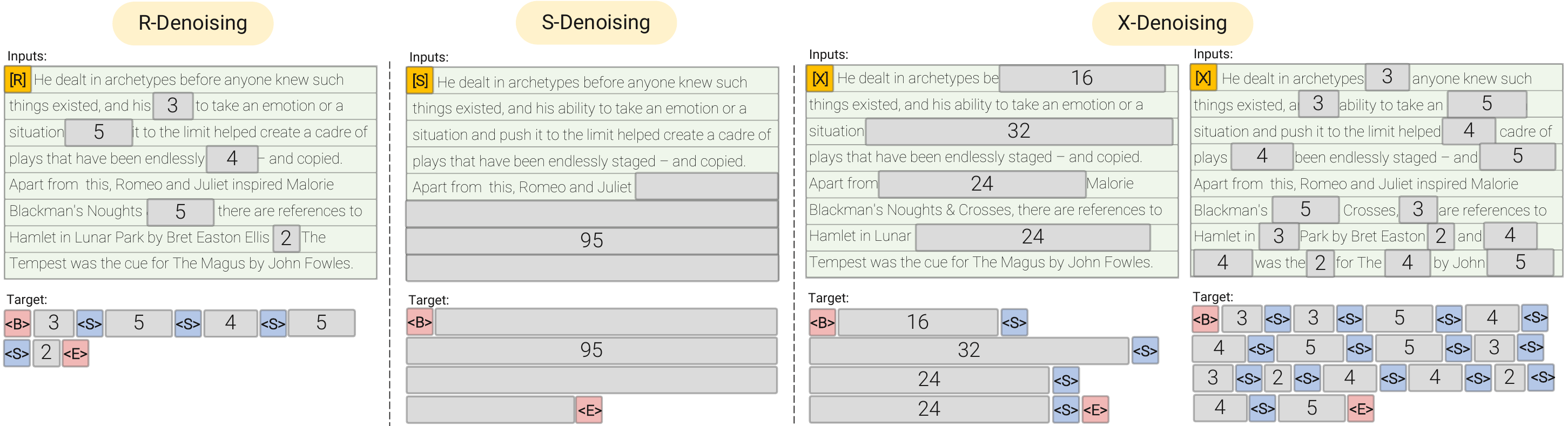}
        \vspace{-1em}
    \label{fig:transformer_arch}
\caption{Mixture of denoisers for training UL2. Greyed out rectangles are masked tokens that are shifted to \textit{`targets'} for prediction.}
\label{stats}
\end{figure*}

\subsubsection{Unified Perspective for Pre-training Tasks}
Many pre-training tasks can be simply formulated as an `input-to-target' task, wherein the input refers to any form of memory or context that the model conditions on, and the target is the model's expected output. Language models use all previous time-steps as inputs to the model to predict the next token, which is the target.  
In span corruption, the model leverages all uncorrupted tokens from the past \emph{and} future as inputs for predicting the corrupted span (targets).
Prefix-LMs are LMs that use past tokens as inputs, but consume the inputs bidirectionally: this offer more modelling power than unidirectional encoding of inputs in vanilla LM.

Given this perspective, we can approximately reduce one pre-training objective to another. For instance, in the span corruption objective, when the corrupted span, i.e., target, is equal to the entire sequence, the problem becomes effectively\footnote{This is roughly approximate since the model still conditions on a sentinel token.} a language modeling problem. With this in mind, using span corruption, by setting the span length to be large, we can effectively mimic the LM objective in local regions. 

We define a notation that covers all of the different denoising tasks that we use in this paper. The inputs and targets of the denoising tasks are generated by a \textsc{SpanCorrupt} function that is parameterized by three values $(\mu, r, n)$, where $\mu$ is the mean span length, $r$ is the corruption rate, and $n$ which is number of corrupted spans. Note that $n$ may be a function of the input length, $L$, and the span length $\mu$, e.g.\ $\nicefrac{L}{\mu}$, but in some cases, we use a fixed value of $n$. Given an input text, \textsc{SpanCorrupt} introduces corruptions to the spans of lengths that are drawn from a (normal or uniform) distribution with mean of $\mu$. After corruption, the input text is then fed to the denoising task and the corrupted spans are used as targets to be recovered. 

As an example, to construct an objective analogous to causal language modeling using this formulation, one would simply set ($\mu=L$, $r=1.0$, $n=1$), i.e. a single span with its span length equal to the length of the sequence. To express one similar to Prefix LM objective, one would set ($\mu=L-P$, $r=1.0 - P/L$, $n=1$) where $P$ is the length of the prefix, with the additional constraint that the single corrupted span always reaches the end of the sequence. 

We note that this inputs-to-targets formulation can be applied to both encoder-decoder models and single-stack transformer models (e.g., decoder models). We opt to select models that predict the next target token instead of those that do so in-place (e.g., predict the current masked token in BERT) because the next-target formulation is more general and can subsume more tasks instead of using a special ``\texttt{CLS}'' tokens and task-specific projection heads.

\subsubsection{Mixture of Denoisers}
We conjecture that a strong universal model has to be exposed to solving diverse set of problems during pre-training. Given that pre-training is done using self-supervision, we argue that such diversity should be injected to the objective of the model, otherwise the model might suffer from lack a certain ability, like long-coherent text generation.  

Motivated by this, as well as current class of objective functions, we define three main paradigms that are used during pre-training:
\begin{itemize}
    \item \textbf{R-Denoiser} - The regular denoising is the standard span corruption introduced in~\citet{raffel2019exploring} that uses a range of $2$ to $5$ tokens as the span length, which masks about $15\%$ of input tokens. These spans are short and potentially useful to acquire knowledge instead of learning to generate fluent text. 
    \item \textbf{S-Denoiser} - A specific case of denoising where we observe a strict sequential order when framing the inputs-to-targets task, i.e., prefix language modeling. To do so, we simply partition the input sequence into two sub-sequences of tokens as context and target such that the targets do not rely on future information. This is unlike standard span corruption where there could be a target token with earlier position than a context token. Note that similar to the Prefix-LM setup, the context (prefix) retains a bidirectional receptive field. We note that S-Denoising with very short memory or no memory is in similar spirit to standard causal language modeling. 
    \item \textbf{X-Denoiser}  - An extreme version of denoising where the model must recover a large part of the input, given a small to moderate part of it. This simulates a situation where a model needs to generate long target from a memory with relatively limited information. To do so, we opt to include examples with aggressive denoising where approximately $50\%$ of the input sequence is masked. This is by increasing the span length and/or corruption rate. We consider a pre-training task to be extreme if it has a long span (e.g., $\geq12$ tokens) \textbf{or} have a large corruption rate (e.g., $\geq30\%$). X-denoising is motivated by being an interpolation between regular span corruption and language model like objectives. 
\end{itemize}

This set of denoisers has strong connections with previously used objective functions: R-Denoising is the T5 span corruption objective, S-Denoising is connected to causal language models that are GPT-like, and X-Denoising can expose the model to a combination of objectives from T5 and Causal LMs. Notably, X-denoisers are also connected to improve sample efficiency since more tokens are learned to be predicted in each sample, in similar spirit to LMs. We propose blending all these tasks in a uniform fashion and have a hybrid self-supervised objective. The final objective is a mixture of 7 denoisers that are configured as follows: 
\begin{table}[H]
    \centering
    \begin{tabular}{cp{8cm}}
    \toprule
    Denoiser & Setting \\
    \midrule
      R   &  $(\mu=3, r=0.15, n) \cup \: (\mu=8, r=0.15, n)$\\
      S   & $(\mu=\nicefrac{L}{4}, r=0.25, 1)$ \\ 
      X &  $(\mu=3, r=0.5, n) \cup \:
            (\mu=8, r=0.5, n) \cup \:
            (\mu=64, r=0.15, n) \cup \:
            (\mu=64, r=0.5, n)$ \\
\bottomrule
    \end{tabular}
    \caption{Configuration of UL2's mixture-of-denoisers used in the paper.}
    \label{tab:my_label}
\end{table}




For X- and R-Denoisers, the span length is sampled from a normal distribution with mean of $\mu$. For S-Denoisers, we use a uniform distribution, fix the number of corrupted spans to 1, and have an additional constraint that the corrupted span should end at the end of the original input text, i.e. no un-cropped token should appear after the corrupted part. This is roughly equivalent to seq2seq denoising or the Prefix LM pre-training objective. 

Since LM is a special case of Prefix-LM, we did not find it necessary to include a casual LM task into the mixture. All tasks have an approximate equal participation in the mixture. We also explore an alternative where we increase number of S-denoisers up to $50\%$ of denoisers in the Mixture and all other denoisers take up the remainder. We present detailed ablation studies of various design choices in the later sections.

Finally, the mixing in Mixture-of-Denoisers is what makes it universally powerful. Alone,  some of the denoiser types do not perform well. For instance, the original T5 paper explored an option with 50\% corruption rate (X-denoising) and found that to not work well. 

The implementation of UL2's mixture of denoiser is simple and easy to implement using a library like seqio\footnote{\url{https://github.com/google/seqio}} \citep{https://doi.org/10.48550/arxiv.2203.17189}. See appendix for more details on implementation.

 
\subsubsection{Mode Switching}
We introduce the notion of paradigm-shifting via mode switching. During pre-training, we feed the model an extra \emph{paradigm} token, i.e., $\{$\texttt{[R]}, \texttt{[S]}, \texttt{[X]}$\}$ that helps the model switch gears and operate on a mode that is more suitable for the given task. 
For fine-tuning and downstream few-shot learning, to trigger the model to learn better solutions, we also add a paradigm token with respect to the setups and requirements of the downstream task. Mode switching in fact binds downstream behavior to one of the modes we used during upstream training. 
\begin{table*}[]
    \centering
    \small
    \caption{Experimental results on a suite of language understanding and generation tasks on both supervised and one-shot setup. Models are pretrained on 32B tokens.}
\label{tab:raw_scores}
    \begin{tabular}{lllccccccccc}
    \toprule
        & & &\multicolumn{4}{c}{Supervised Finetuning} & \multicolumn{4}{c}{In-context One-shot}\\
        Obj &  Arch &  Params &SG & XS & SGD & TOT & SG & XS & SGD & TOT   & LM  \\
        \midrule
  CLM       & Dec & 167M & 62.24 & 	28.18& 	55.44 & 	59.40 & 	39.22	&  1.16 & 	1.40 & 	0.20 & 	-2.35 \\ 
  PLM & Dec & 167M  & 62.44 &	28.21 &	55.55 &	59.52 &	42.54 &	1.08 & 	3.70	& 6.40 &	-2.54\\
  SC & Dec & 167M  &67.67 &	29.14	& 55.48 	& 	60.47 	& 	38.53 	& 	1.16 	& 	2.20 	& 	1.60 	& 	-3.62 \\ 
  SCLM & Dec & 167M & 63.36 &	29.02 &	55.71 &	60.00 &	40.78 &	3.03 &	1.27 &	0.10 &	-2.38	 \\
  UL2 & Dec & 167M &  65.50	& 28.90	 &  55.80 & 	60.39 & 	\textbf{42.30} & 	8.01 & 	6.30 & 	5.80 & 	\textbf{-2.34} \\
  \midrule 
  PLM & ED & 335M & 69.30	& \textbf{31.95} & 	55.70& 	60.91& 	38.18& 	6.50& 	7.11& 	3.90& 	-2.42 \\ 
  SC & ED & 335M &  72.00	&  31.05 & 	55.80 & 	61.25 & 	38.51 & 	7.49 & 	1.43 & 	2.10 & 	-7.23 \\
  SCLM & ED &  335M & 72.50	 & 31.69 &	55.70 &	60.94 &	39.74 &	5.13&	\textbf{8.70}&	\textbf{7.30}&	-2.40 \\
  UniLM & ED &  335M &  71.10 &	31.00 &	55.83 &	61.03 &	39.86 &	6.70	&6.50&	4.10&	-2.65 \\
  UL2 & ED & 335M &  \textbf{73.10} &	31.86&	\textbf{56.10}&	\textbf{61.50}&	41.30&	\textbf{11.51}&	6.63	&6.50	& -2.55\\ 
  \bottomrule 
    \end{tabular}
    
\end{table*}
\subsection{Model Architecture}
UL2 adopts an architecture-agnostic philosophy. We argue that the choice between both architectures (encoder-decoder vs decoder-only) is a more of an efficiency trade-off and that architecture choice should not be conflated with the pretraining objective. Hence, we have both a UL2 decoder and UL2 encoder-decoder in similar spirit to how there are multiple sizes per model. We discuss this efficiency trade-off in detail in our experiment section. UL2 adopts a pretty standard vanilla T5 Transformer that have been enhanced with modifications that have withstood the test of time, i.e., GLU layers \citep{shazeer2020glu} and T5-style relative attention. To not further conflate architectural modifications with pretraining contributions, the backbone of the model remains similar to a T5-like model. This is also in light of results such as \citep{narang2021transformer}. 
 

\section{Ablative Experiments}
This section describes our ablative experimental setup (e.g., baselines, datasets, implementation details) and results. Our overall findings show that UL2 outperforms T5-like and GPT-like models on 9 out of 9 tasks.
\subsection{Baselines}
For pre-training objectives, we compare with the following pre-training baselines:
\begin{itemize}
    \item \textbf{Causal Language Model (CLM)} - This is the standard left-to-right auto-regressive language model pre-training, used in many standard pre-trained models, like GPT~\citep{Radford2019,brown2020language}. We refer to this model as GPT-like in our experiments.
    \item \textbf{Prefix LM (PLM)} - This is a slight variation of causal LM where $M$ has bidirectional receptive fields, introduced in \citep{liu2018generating,raffel2019exploring}. We uniformly sample the length of $M$ and only compute the loss at the auto-regressive targets. 
    \item \textbf{Span Corruption (SC)} - This is the standard denoising objective proposed in T5 \citep{raffel2019exploring}. The idea is to blank out certain text portions and replace them with sentinel tokens. The text replaced with sentinel tokens are then copied to the targets and auto-regressively generated by the model. We use a mean span of $3$ and denoising rate of $15\%$ following the default T5 setup.  
    \item \textbf{Span Corruption + LM (SCLM)} - We train on a mixture of CLM and Span Corruption with an equal mix ratio. We use the same hyper-parameters for SC for the SC component of this objective.
    \item \textbf{UniLM (ULM)} - This is the objective proposed in \citet{dong2019unified}. Similar to the original UniLM, we mix causal language modeling, Prefix LM (sequence-to-sequence LM) and bidirectional i.i.d denoising. Instead of training UniLM in cloze-style or BERT-style, we opt to generate the masked tokens.  This allows this objective to be applicable to both decoder-only and encoder-decoder architectures and remove the need for task-specific linear heads for fine-tuning. 
\end{itemize}
For all objectives, we explore both single-stack and encoder-decoder architectures. All architectures are inputs-to-targets either implemented in encoder-decoder or decoder-only model structures since we consider BERT-style masked language modeling pretraining to have already been effectively subsumed by this style of pretraining, as empirically made evident in \citep{raffel2019exploring}. Task-specific classification heads are also not recommended, since they clearly go against the principle of having a universal model (and are also very cumbersome).


\subsection{Experimental Setup}
We conduct our experiments on a diverse set of supervised and prompt-based few-shot learning tasks. 

\subsubsection{Datasets and Tasks}
The datasets we use are SuperGLUE \citep{wang2019superglue}, comprising of 8 NLU sub-tasks. We also conduct experiments on 3 datasets from the GEM benchmark \citep{gehrmann2021gem} that focuses on language generation problems. We arbitrarily select XSUM (summarization), ToTTo (table-to-text generation) \citep{parikh2020totto} and Schema Guided Dialog (SGD) \citep{rastogi2019towards} from the GEM benchmark. For all these tasks, we evaluate on both supervised fine-tuning and prompt-based one-shot learning. Finally we also compare our models on their general ability for text generation using perplexity scores on the C4 validation set. We believe our suite of tasks gives us good coverage across many setups in the literature including supervised and conditional few-shot learning.

\subsubsection{Metrics and Holistic Evaluation} For SuperGLUE, we report well-established metrics such as accuracy, F1 or Exact Match, whenever appropriate. For GEM benchmark, we use the Rouge-L metric. For language modeling we report negative log perplexity.
The \emph{universality} of the models, i.e., their collective performance across all range of tasks, is a main evaluation criteria here. To enable the comparison between models from this perspective, we need an aggregate performance score. However, metrics on different tasks we include are widely different in nature -- take, for example, F1 and perplexity. To address this, we opt to report and use the \textit{normalized relative gain with respect to baselines} as an overall metric. For this purpose, we use the standard language model (decoder-only) (GPT-like) and standard span denoising encoder-decoder (T5) as prime baselines and report all methods against their relative performance against these well-established candidates. We believe this is the most suitable method for comparing these models since it is easy to reason about how much a new model is generally better than a popular setting (e.g., GPT or T5-like). We also highlight the fact that the overall gain is \textbf{normalized}, so this becomes harder to exploit or be susceptible to benchmark lottery effects \citep{dehghani2021benchmark}. 

\subsubsection{Implementation Details}
Our experiments are all conducted in JAX/Flax \citep{jax2018github} using the open source T5X\footnote{\url{https://github.com/google-research/t5x}.} framework \citep{https://doi.org/10.48550/arxiv.2203.17189} and Flaxformer\footnote{\url{https://github.com/google/flaxformer}}. We pre-train all models for 500K steps with a batch size of 128 and a sequence length of 512 inputs and 512 targets using the C4 corpus. The total approximate tokens seen during pre-training is approximately 32 billion tokens. Each pre-training run is typically trained using 64 to 128 TPUv4 chips \citep{jouppi2020domain}. We optimize our model with the Adafactor \citep{shazeer2018adafactor} optimizer with an inverse square root learning rate. To understand the trade-off of different backbone architectures, we run all baseline pre-training objectives with both the decoder-only architecture and encoder-decoder architecture. We report key experiment results using a base architecture of approximately 167M parameters for the decoder model and 335M parameters for the encoder-decoder model. All models use a standard Transformer that uses SwiGLU layers as described in \citep{shazeer2020glu}. We utilize the default T5 English 32K sentencepiece for all models. Within the context of decoder-only models, except for the case of the decoder model trained on causal LM, our experiments always use a bidirectional receptive field \textbf{only} in it's input segment and autoregressive decoding at the \textit{targets} segment. This is essentially the a PrefixLM-type architecture\footnote{Not to be confused with the PrefixLM pretraining objective.} \citep{raffel2019exploring} which we find to be consistently better than a full causal decoder model.

\begin{table*}[]
    \centering
\small
    \caption{Relative performance compared to standard encoder-decoder span corruption model (T5). Results in this table are expressed in terms of relative percentage improvements over a baseline. Model with $\star$ denotes the main compared baseline. Overall score column is normalized to be weighted equally across tasks.}
 \label{tab:relative_t5}
\begin{tabular}{llccccccccccc}
    \toprule
    & & \multicolumn{4}{c}{Supervised} & \multicolumn{4}{c}{One-shot}\\
        Obj &  Arch & SG & XS & SGD & TOT & SGL & XS & SGD & TOT   & LM & All & Win \\
        \midrule
  CLM       & Dec & -13.6 &	-9.2&	-0.7&	-3.0&	+1.8&	-91.7&	-2.2&	-90.5&	+208 & -31.7 & 2/9\\ 
PLM & Dec & -13.3	&  -9.2 & 	-0.5 & 	-2.8 & 	+10.5 & 	-85.6 & 	+158 & 	+205 & 	+185 &-11.0 & 4/9\\
SC & Dec & -5.6&	-6.2 &	-0.6&	-1.3&	+0.05&	-84.5&	+54&	-23.8&	+99 & -20.6 & 3/9\\
SCLM & Dec & -6.0	& -6.5 &	-0.2 &	-2.0  &	+5.9 &	-59.6	& -11.3 & 	-95 &	+204 & -16.1  & 2/9 \\
UniLM & Dec & -10.1 & 	-8.2 &	-0.2 &	-2.3 &	-5.3 &	-69.1 &	+382 &	+110 & 	+200 & -16.1 & 3/9\\
UL2 & Dec & -9.0 &	-6.9&	0.0&	-1.4&	+9.8&	+6.9&	+340&	+176	& +209 &  \textbf{+14.1} & 5/9 \\
  \midrule 
PLM & ED & -3.7& 	+2.9 &	-0.2&	-0.6	&-0.86&	-13.3&	+397	&+86&	+199 & +16.7 & 5/9 \\  
 SC$^\star$ & ED & 0.0	& 0.0	 & 0.0 &	0.0&	0.0 &	0.0 & 0.0 &	0.0 &	0.0 &	0.0 & - \\
 SCLM & ED & +0.7 &	+2.1	& -0.2 	& -0.5  &	+3.2 &	-31.6 &	+508 &	+248 &	+201 & +28.3 & 7/9 \\
 UniLM & ED &  -1.2 &	-0.2 &	+0.1 &	-0.4 &	+3.5 &	-11.0 &	+355 &	+95	 &+173 & +19.8 & 5/9\\
UL2 & ED & +1.5	& +2.6	&  +0.5 & 	+0.4 & 	+7.2& 	+53.6& 	+363& 	+210& 	+184 & \textbf{+43.6} & \textbf{9/9}\\
  \bottomrule 
    \end{tabular}
    
\end{table*}

\begin{table*}[h]
    \centering
\small
    \caption{Relative performance compared to standard decoder causal language model (GPT-like). Results in this table are expressed in terms of relative percentage improvements over a baseline. Model with $\star$ denotes the main compared baseline. Overall score column is normalized to be weighted equally across tasks.}
     \resizebox{\textwidth}{!}{\label{tab:relative_gpt}\begin{tabular}{llccccccccccc}
    \toprule    & & \multicolumn{4}{c}{Supervised} & \multicolumn{4}{c}{One-shot}\\
        Obj &  Arch & SG & XS & SGD & TOT & SG & XS & SGD & TOT   & LM & All  & Win \\
        \midrule
  CLM$^\star$       & Dec & 0.0 &	0.0  &	0.0  &	0.0  &	0.0  & 	0.0  &	0.0  & 	0.0  &	0.0  & 0.0 & -\\ 
PLM & Dec & +0.3 &	+0.1	& +0.2	&+0.2 & 	+8.5	& +74.3 &	+164 &	+3100	 & -8.0  & +21.4 & 8/9 \\
UniLM & Dec & +4.0 & 	+1.1 &	+0.5& 	+0.7	&-7.0 &	+274	& +393 &+2100&	-2.5 & +21.0 & 7/9 \\
SC & Dec & +8.7	& +3.4 & 	+0.1 &	+1.8 &	-1.8	& +87.0 &	+57.1 &	+700 &	-54.2 & +13.9 & 7/9\\
SCLM & Dec & +1.8	& +3.0&	+0.5 & 	+1.0 &	+4.0	& +387 &	-9.3 & 	-50 &	-1.3 &+15.8 & 6/9 \\
UL2 & Dec & +5.2	& +2.6&	+0.6&	+1.7 &	+7.9 &	+1190	& +350 &	+2800	& +0.3 & \textbf{+45.7} & \textbf{9/9} \\
  \midrule 
PLM & ED & +11.3 &	+13.4  &+0.5 &	+2.5 &	-2.6	& +946&	+408& 	+1850	& -2.9 & +48.6  & 7/9 \\  
 SC & ED & +16.5 &	+10.2&	+0.6&	+3.1&	-1.8&	+1107	&+2.3	&+950	 &-208 & +31.7 & 7/9   \\
 SCLM & ED & +15.7 &	+12.5 & +0.5 &	+2.6 & +1.3& 	+726 &	+522 &  +3550 &	-2.2  & +60.3 & 8/9\\
 UniLM & ED & +14.2 & +10.0 & 	+0.7 &+2.7 &+1.6 &+974	& +365 & 	+1950 &	-12.9 &  +52.6 & 8/9\\
UL2 & ED & +17.4 &	+13.1	& +1.2 &	+3.5 & 	+5.3 & 	+1754	& +373 &	+3150 &	-8.3  & +\textbf{76.1} & 8/9\\
  \bottomrule 
    \end{tabular}}
\end{table*}
\subsection{Overview of Ablative Experimental Results}
Table \ref{tab:raw_scores} reports the raw results on all the benchmark tasks and datasets. To facilitate easier comparison across setups, we also report relative comparisons against well-established baselines such as T5 and GPT models. This is reported in Tables \ref{tab:relative_t5} and \ref{tab:relative_gpt} respectively. 
\subsubsection{Decoder Vs Encoder-Decoder}
Before we dive into the results of this segment, we would like to remind readers that there is no easy way to compare decoder-only models with encoder-decoder models. In short, we can either compare them in a compute-matched setup or a parameter-matched way. Therefore, the encoder-decoder models in these set of results have approximately twice the number of parameters as the decoder models but have similar speeds. 

We note that this may slightly favor encoder-decoders since this can be interpreted form of model sparsity. Moving back to the results, when using T5 as the reference baseline, we note that, with the exception of UL2 Decoder, none of the pre-trained decoders models outperform T5. Additionally, there is a $10\%$ to $30\%$ degradation in overall relative performance. The best decoder baseline model here is the Prefix-LM decoder model, which is about $10\%$ worse than the T5 baseline. It is clear from these results that encoder-decoder models should be preferred over decoder-only models if and only if there is no concern about storage, i.e., parameter counts are generally less important than actual throughput (see \citep{dehghani2021efficiency} for a detailed discussion). 

When there is a parameter constraint, the Prefix-LM decoder makes for a suitable alternative. Finally, an interesting data point is how we were able to push the UL2 decoder to outperform the T5 encoder-decoder setup by $+14.6\%$. That said, this UL2 decoder does not outperform our UL2 encoder-decoder. However, this reinforces our point that the self-supervision objective may be intrinsically more important than the backbone architecture and negotiating architectural choices is mainly about efficiency trade-offs that can be studied independently.

\subsubsection{Is GPT and/or T5 the optimal setup?}
Based on the relative comparisons against a GPT-like (causal LM + decoder) and T5-like (span corruption + encoder decoder) setup, we are able to easily identify if the well-established setups are indeed optimal or already close to optimal. Firstly, the causal LM (GPT-like) setup appears to be the worse configuration as it is outperformed by all our baselines. We thus make the straightforward recommendation of always at least training with Prefix-LM or UniLM whenever possible. The best decoder-only model (with the exception of UL2) is the Prefix-LM pre-training that keeps a memory prefix for a language model to condition on. Regarding Prefix-LM pre-training, it is interesting that Prefix-LM actually outperforms the T5 span corrupt setup by $+16.7\%$. The Prefix-LM encoder-decoder model is indeed less effective than the default T5 model on SuperGLUE but is on a whole, stronger especially when it comes to one-shot or open text-generation. Overall, between the Prefix-LM and the span corruption encoder-decoder model (T5), it is unclear to which is the universally superior model as there are gives and takes across the different sub-tasks although it is worthy noting the Prefix-LM EncDec model only sacrifices a minor degradation in certain tasks for a huge multifold increase in other tasks. 

\subsubsection{On the Performance of UniLM and SCLM}
On the encoder-decoder setup, both the UniLM and SCLM objective performs better than the standard span corruption objective in terms of aggregated and normalized overall gain. This shows that, in general, mixing pre-training objectives is helpful. On the decoder setup, there is an overall gain of $+9.4\%$ gain for UniLM and $+16.1\%$ for SCLM compared to the baseline causal LM. In terms of individual tasks, UniLM and SCLM both outperforms T5 on 6 out of 9 tasks. It is also noteworthy that SCLM performs the best out of all models on 1shot generation (SGD and TOTTO). 

\subsubsection{On the Performance of the Proposed UL2}
Finally, we note that UL2 performs the best when compared against both the GPT-like model and the T5-like model. Overall, UL2 outperforms by T5 $+43.4\%$ and $+76.2\%$ when compared to the GPT-like CLM decoder model. This is the highest relative (overall) gain compared to all other alternatives. We also note that on all individual tasks, UL2 outperforms T5 on \textbf{all 9 out of 9} considered tasks. Hence, UL2 is a universally better option compared to the span corruption T5 model. While UL2 doesn't always outperform all baselines on all individual tasks, UL2 is very consistent. Even when it loses to another method on a task, the loss is relatively marginal (e.g., 6.5 vs 7.3 on one-shot TOTTO). Conversely, when UL2 outperforms a baseline like T5, the gain can be as large as $+363\%$. UL2 remains the most consistently strong method. The consistent improvement also suggests that it can be used as a more consistent replacement to T5 and GPT-like models.

\subsection{Mode Switching Ablations}
In order to ascertain that our mode switching capabilities have an effective on performance, we conduct ablation experiments.
\begin{table}[t]
    \centering
     \caption{Effect of different paradigm prompts on 1-shot evaluation, using a Encoder-Decoder architecture pre-trained using UL2  on 7B tokens.}
    \label{tab:prompt_effect}
    \begin{tabular}{lcc}
    \toprule
       Model/Prompt  & 1Shot XSum & 1Shot SuperGLUE \\
         \midrule
         Baseline T5 &  6.9/0.6/6.1 & 33.9 \\
       UL2 / None & 13.2/1.4/10.8 & 38.3 \\
       UL2 / \texttt{[R]}  & \textbf{13.5}/\textbf{1.5}/\textbf{11.1} & 38.5 \\
       UL2 / \texttt{[S]} &11.6/1.2/10.0& 38.5\\ 
       UL2 / \texttt{[X]} & 8.9/0.9/7.6 & \textbf{38.7}\\
         \bottomrule
    \end{tabular}
\end{table}
We conduct experiments on one-shot XSum and one-shot SuperGLUE. Table \ref{tab:prompt_effect} reports the result of varying the paradigm prompt to the model. Firstly, we observe that the prompt has quite substantial effect on model performance -- i.e., using the right or wrong prompt can lead to a 48\% gap in performance (on XSum, Rouge-1). SuperGLUE, on the other hand, is less sensitive to prompting. On SuperGLUE, using prompts are almost always better than not using prompts during one-shot eval. However, for XSum, getting the prompt right seems to be crucial for good performance.


  

\subsection{Mixture-of-Denoisers Ablations}
\begin{table}[t]
    \centering
    \small
      \caption{Ablation study for Mixture-of-Denoisers. Span, Rate and SD are in percentages ($\%$). We report SuperGLUE score (SG) and XSUM Rouge-L (XS).}
    \label{tab:denoiser_ablation}\begin{tabular}{llllcccc}
    \toprule
       & \multicolumn{3}{c}{Ablation Method} & \multicolumn{2}{c}{Supervised}  &  \multicolumn{2}{c}{One-shot}  \\
        Name & Span ($\mu$) & Rate ($r)$ & SD\% & SG & XS & SG & XS \\
        \midrule
       A &  - & - & 100  & 69.3 & 31.1 & 38.2 & 6.5\\
        B & 3 & 50 & 0  &72.0 & 32.0 & 38.5 & 7.5\\
        \midrule
C &   3,8,12 & 15,50    & 14 & 71.9 & 32.1 & 38.6 & 4.1\\
 D &3,8,12,32 & 15,50 &  11 &71.0 & \textbf{32.2} & \textbf{42.7} & 10.6\\
 E & 3,8,32,64 & 15,50 & 11 &  73.1 & \textbf{32.2} & 40.7 & 10.4\\
 F & 3,8,64 & 15,50 & 17 & 70.6 & 31.6 & 41.3 & 11.5 \\
 G & 3,8,32,64 & 15 & 25 & 69.2 & 31.6 & 42.4 & 10.1\\ 
 H  & 8, 64 & 15 & 25 & 72.5 & 31.2 & 39.2 & 10.9 \\
  I & 3,8,12, 32 & 15,50 & 50 & 71.2 &32.0 & 38.1 & 11.7\\
  J & 3,8,64 & 15,50 & 50 & 71.3  &31.6 & 38.1& \textbf{11.8}\\
  K & 3,8,12 & 15,50 & 0 & \textbf{73.7}  & 32.0& 39.3 & 2.6\\
  L & 3,8,64 & 15,50 & 0 & 70.1 & 32.1& 38.0& 7.3\\
         \bottomrule
    \end{tabular}
  
\end{table}
We conduct extensive experiments to verify the effectiveness of individual objectives within the MoD objective. Table \ref{tab:denoiser_ablation} reports results for these ablations. We report results for varying the mean span, and corruption rate, along with the percentage of S-denoising used (denoted by \% SD)). Note that the total number of denoisers in a mixture is $\|Span\| \times \|Corrupt\_Rate\| + 1$. 
We label these configurations from Var-A through Var-J to refer to them easily.

\paragraph{X-Denoising is Complementarily Effective but Does Not Suffice as a Standalone}
We observe that mixing Extreme Denoising is effective. Most of the best results across the board come from mixtures with long spans (e.g., 32 or 64). When compared with variants without long spans (Var-D vs. Var-C), we see that Var-D is strictly better. We also draw the readers attention to Var-H, which is a variant that only employs long spans. In general, Var-H performs poorly, suggesting that extreme denoising complements regular denoising but does not suffice in isolation. This also corroborates the result from \cite{raffel2019exploring} that shows that a $50\%$ corruption rate does not perform well. This slightly conflicts with the finding of \citep{wettig2022should} although our architectures use a inputs-to-targets form of pretraining instead of BERT-style masked language modeling. 

\paragraph{Small Amounts of S-Denoisers is Preferred} We explore a setting where we scale S-denoisers to $50\%$ of the entire MoD mixture. We find that this generally hurts performance. Hence, we make a conclusion that S-denoisers are necessary but only small amounts of S-denoisers ($\approx 20\%$) are preferred. Var-K and Var-L also explore the case where there is no S-denoising at all. While performance on one task substantially improves (SuperGLUE), another substantially degrades (one-shot XSUM). Meanwhile for Var-L which is identical to var-F (but without S-denoising), performs on a whole, substantially worse. Hence, we showed that S-denoising is crucial.

\subsection{Modestly Scaling Model Size and Pretraining Data}
We conduct additional experiments by scaling up both 1) the model size and 2) pre-training dataset size. Concretely, we scale the UL2 Encoder-Decoder model up to approximately 1B parameters and increase the number of pre-training tokens to 0.5 trillion tokens. Our motivation is to conduct a sanity check that the proposed formulation also works at different model scales and to observe if there are differences and implications at operating at a larger scale. Moreover, it has also become a staple for language model research to derive scaling laws \citep{kaplan2020scaling,tay2021scale}. Table \ref{tab:scale} reports results in this scaled setting. At large scale, we find that the proposed of the UL2 encoder-decoder model is still competitive. A key difference now is that UL2 drops the SuperGLUE suite against T5 (1B). However, this is compensated by not only out-performing on 7 out of 8 tasks but also improving performance by 2-4 times on one-shot evaluation. The gains on supervised fine-tuning is smaller, but still noticeable across the board on XSUM, SGD and TOT.

\begin{table}[H]
    \centering
     \caption{Experiments with moderately scaled up models in terms of model compute (e.g., 1B for EncDec and 0.5B for decoder-only) and dataset size (0.5T tokens).}
    \label{tab:scale}
  \begin{tabular}{lcccccccc}
    \toprule
    & \multicolumn{4}{c}{Finetuning} & \multicolumn{4}{c}{In-context Learning}\\
    Model  & SG & XS & SGD & TOT & SG & XS & SGD & TOT  \\
    \midrule
    GPT-like  & 62.3 &37.1/15.7/30.2& 56.0& 60.3 & 36.4 & 1.2/0.1/1.1 & 3.5 & 0.0\\ 
        T5   & \textbf{84.7} & 43.0/20.8/35.6 & 56.0 & 62.1&29.4 & 8.9/0.8/7.8  & 2.1 & 1.4 \\
        UL2  & 83.3 & \textbf{43.3/21.0/35.9}  & \textbf{56.5}& \textbf{62.6} & \textbf{45.4} & \textbf{15.4/2.5/11.1} & \textbf{9.6} & \textbf{7.8} \\
        \bottomrule
    \end{tabular}
\end{table}

\section{Scaling to 20B Parameters} 
We are also interested to evaluate UL2 in a scaled up setting. Following the insights we obtained from ablative experiments, we use an encoder-decoder architecture for this run. While UL2 is architecture agnostic, our \textit{soft} advice here is to probably default to an encoder-decoder architecture due to intrinsic sparsity. 

We train UL2 at a scale of approximately 20B total parameters. Compared to truly large language models \citep{du2021glam,chowdhery2022palm}, 20B represents a medium scale model that we train as a proof-of-concept resembling a hint of what UL2 can do at a relatively larger scale than our ablation experiments. Admittedly, not much thought was put into the exact parameter count of this model, i.e., we were training a 20B model already for some time and decided to see it to convergence. Additionally, we note that spiking and instabilities are common when scaling up models due to a potential barrage of reasons (data corruption, intermittent hardware issues like pre-emption). In this run we did not specifically control or put in place any mitigation strategies such as occasional restarts as we were not attentively monitoring the job. Hence, we find occasional loss spikes in the training of this 20B model. However, since many finetuning experiments using these checkpoints still often result in \emph{sota} performance, we let it be for now and leave a properly monitored run for future work. Despite obtaining \textit{sota} performance on 50+ NLP benchmarks, we expect the current presented results to be still an underestimate of the true potential of the model. We leave properly scaling UL2 to truly large scale to future work.

\subsection{Pretraining and Model Configuration}
We follow the same training protocol in earlier experiments by pretraining on the C4 corpus but by also scaling the number of tokens the model sees during pretraining. We use a batch size of 1024 and 512 TPUv4 chips for pretraining this model. The model is trained on a total of 1 trillion tokens on C4 (2 million steps). The sequence length is set to $512/512$ for inputs and targets. Dropout is set to $0$ during pretraining. Pre-training took approximately slight more than one month for about 1 trillion tokens. We use the same mixture of denoisers as earlier sections. The model has $32$ encoder layers and $32$ decoder layers, $d_{model}$ of 4096 and $d_{ff}$ of $16384$. The dimension of each head is $256$ for a total of $16$ heads. Our model uses a model parallelism of $8$. We retain the same sentencepiece tokenizer as T5 of 32k vocab size. Hence, UL20B can be interpreted as a model that is quite similar to T5 but trained with a different objective and slightly different scaling knobs. Similar to earlier experiments, UL20B is trained with Jax and T5X infrastructure. We release and open source T5X-based model checkpoints of this 20B model.

\subsection{Experiments at 20B scale}
This section describes our experimental setup for UL20B experiments.
\subsubsection{Setup and Implementation Details}
We conduct experiments on both finetuning and in-context learning. For supervised finetuning, our models are continuously finetuned after $N$ pretraining steps where $N$ is typically from $50k$ to $100k$. In other words, after each $N$k steps of pretraining, we finetune on each downstream task and record its results. This is generally done in a manual fashion. While some tasks were finetuned on earlier pretrained checkpoints as the model was still pretraining, many were finetuned on checkpoints nearer to convergence that we release. As we continiously finetune, we stop finetuning on a task once it has reached \textit{sota} to save compute. Finetuning is generally done in a per-task basis and not co-trained. Details of tasks where co-training is performed is found in the appendix. We leave the combination of massive multi-task training \citep{aribandi2021ext5} and UL2 to future work. 

For supervised finetuning, we generally adopt a learning rate in the range of $\{5 \times 10^{-5}, 1 \times 10^{-5}\,  1 \times 10^{-4}\}$ using the Adafactor optimizer. The general recipe is that we reset Adafactor optimizer states and/or adopt a loss normalization based on the number of real target tokens. This is reminiscent of the PaLM finetuning setup \citep{chowdhery2022palm}. Batch size is generally a range of $32$ to $128$ although we did not find much impact of batch size on finetuning performance. Many of the evaluated tasks were not tuned much and we only ran once or twice before performing leaderboard submissions.  

\subsubsection{Datasets for Supervised Finetuning}
To demonstrate the universality of the approach, we consider a total of nearly 50+ NLP tasks. We list our categorization of tasks below. Note that the categorization of tasks are generally soft in nature and some tasks may cross into different categorization boundaries.

\begin{itemize}
    \item  \textbf{Language Generation} - We consider summarization and data-to-text generation tasks. We use CNN/Dailymail \citep{hermann2015teaching}, XSUM \citep{Narayan2018DontGM}, MultiNews \citep{fabbri2019multi}, SAMSum \citep{gliwa2019samsum}, WebNLG \citep{castro-ferreira20:bilin-bi-direc-webnl-shared} (English), E2E  \citep{e2e_cleaned} and CommonGen \citep{lin-etal-2020-commongen} to evaluate our models. For WebNLG, E2E and CommonGen, we use the versions from the GEM benchmark \citep{gehrmann2021gem}. 
    \item \textbf{Language Generation with Human Evaluation} - We evaluate on a variety of text generation tasks using human evaluation, via the GENIE leaderboard \citep{genie}. These tasks include aNLG \citep{bhagavatula2019abductive}, ARC-DA \citep{ARC}, WMT19 \citep{wmt19translate}, and XSUM \citep{Narayan2018DontGM}. 
    \item \textbf{Language Understanding, Classification and Question Answering} - We use Reading Comprehension, Question Answering, Text Classification and natural language inference datasets. Concretely, we use RACE (Reading comprehension) \citep{lai-etal-2017-race}, QASC \citep{khot2020qasc}, OpenBookQA \citep{mihaylov2018can}, TweetQA \citep{xiong2019tweetqa}, QuAIL \citep{rogers2020getting}, IMDB \citep{maas-EtAl:2011:ACL-HLT2011}, Agnews \citep{zhang2015character}, DocNLI \citep{yin2021docnli}, Adversarial NLI \citep{nie2019adversarial}, VitaminC \citep{schuster2021get}, Civil Comments and Wikipedia Toxicity detection datasets \citep{DBLP:journals/corr/abs-1903-04561}. We also use standard SuperGLUE \citep{wang2019superglue} and GLUE \citep{wang2018glue} datasets.
    \item \textbf{Commonsense Reasoning} - We use HellaSwag \citep{zellers2019hellaswag}, SocialIQA/SIQA \citep{sap2019socialiqa}, PhysicalIQA/PIQA \citep{bisk2020piqa}, CosmosQA \citep{huang2019cosmos}, AbductiveNLI \citep{bhagavatula2019abductive}, CommonsenseQA \citep{talmor2018commonsenseqa}, CommonsenseQA2 \citep{talmor2021commonsenseqa}.
    \item \textbf{Long Range Reasoning} - We use the Scrolls benchmark \citep{shaham2022scrolls} which comprises of seven component tasks including GovReport \citep{huang2021efficient}, SumScr \citep{chen2021summscreen}, QMSUm \citep{zhong2021qmsum}, QASPER \citep{dasigi2021dataset}, NarrativeQA \citep{Kocisky2018narrativeqa}, QuaLITY \citep{pang2021quality}, and ContractNLI \citep{koreeda2021contractnli}.
    \item \textbf{Structured Knowledge Grounding} - We use several component tasks from UnifiedSKG \citep{xie2022unifiedskg}, namely WikiTQ \citep{pasupat2015compositional}, CompWQ \citep{talmor2018web}, FetaQA \citep{nan2021fetaqa}, HybridQA \citep{chen2020hybridqa}, WikiSQL \citep{zhongSeq2SQL2017}, TabFat \citep{chen2019tabfact}, Feverous \citep{aly-etal-2021-fact}, SQA \citep{iyyer2017search},  MTOP \citep{li2020mtop} and DART \citep{nan2020dart}. We select datasets that are relatively convenient to perform evaluation and uses mainstream metrics such as accuracy or exact match instead of obscure ones or those that require significant domain specific post-processing. 
    \item \textbf{Information Retrieval} - IR is the task of retrieving relevant documents given queries. We use the setup of the latest next generation IR paradigm, i.e., differentiable search index \citep{tay2022transformer} for our experiments. We use the same NQ \citep{Kwiatkowski2019naturalquestions} splits in the DSI paper.
\end{itemize}
For each dataset, we report the best previous sota result. For generation tasks, we generally report ROUGE-2 following the advice of \citep{gehrmann2022repairing}. For the rest of the datasets, we report the dominant metric that is reported in prior work. For BLEU scores, we use sacrebleu. For commonsense reasoning tasks, we do not compare against approaches that use external knowledge bases as they are orthogonal and out of scope for this paper. For most part, GLUE is generally considered to be saturated and there are many unpublished results on the GLUE leaderboard. For this reason, we make a very reasonable decision of considering \citep{raffel2019exploring} to be the state-of-the-art since we believe that there has not been any real advance on the GLUE benchmark since the T5 model \citep{raffel2019exploring}. GLUE results, given how saturated it already is, are provided as a reference and should be taken with a pinch of salt. 

Generally, we make a best effort to submit scores to any leaderboard (unpublished test set) but refrain from doing so in the cases where the labor costs to make such a submission is prohibitive - especially when the existing state-of-the-art approach has made their dev scores available or when reporting on this particular dataset is only for completeness (e.g., GLUE). We would advise readers to not over think the differences in dev/test since (1) in most academic leaderboards, dev/test aligns not only from our own experience but also can be empirically observed and because (2) the real test is production anyways. Whenever reporting on leaderboard, we consider the top performing published work as SOTA and indicate in our results using the $\#$ symbol that there might be some anonymous submission that has scored higher. For this purpose we consider arxiv preprints of above reasonable quality to count as published work. These results and comparisons are accurate as of 15th April 2022 where we stopped experiments to focus on polishing this paper. We later realized, while preparing to put this paper up on arxiv that there have been new results on Scrolls benchmark using a model \citep{guo2021longt5} using 16k sequence lengths as opposed to ours (2k) where we kept it at 2k once we had obtained sota. It is expected that increasing the length to UL2 would significantly improve our scores likely above current sota, but in the interest of logistics and timeline we leave that to future work.

\subsubsection{Summary of Supervised Finetuning Results}
This section describes the overview results of our experiments.
\setlength{\LTcapwidth}{\linewidth}
\begin{center}
\small
\begin{longtable}{lllllll}

     \caption{Summary of UL20B results compared to state-of-the-art. $(l)$ denotes leaderboard submission. $(^\sharp)$ denotes the best published we could find on the leaderboard. $(e)$ denotes SOTA used an ensembled approach. Because we evaluate finetuning and in-context trade-offs for SuperGLUE, SuperGLUE scores have their own dedicated section below.} \\
         \label{tab:big_table} \\
\toprule
    \textbf{Dataset}     & \textbf{Metric} & \textbf{Eval} & \textbf{Sota Reference}  &  \textbf{SOTA} & \textbf{Ours} \\
    \midrule
    \endfirsthead
    
    \multicolumn{6}{c}%
{{\bfseries \tablename\ \thetable{} -- continued from previous page}} \\
\midrule \\
\endhead

\midrule \multicolumn{6}{r}{{Continued on next page}} \\
\endfoot

\bottomrule
\endlastfoot

       CNN/DM  & Rouge-2    & Test & \citeauthor{zoph2022designing}  &   21.7 & \textbf{21.9}    \\
    XSUM  & Rouge-2 & Test & \citeauthor{zoph2022designing} &\textbf{27.1} & 26.6 \\ 
    MultiNews & Rouge-2 & Test & \citeauthor{xiao2021primer}&21.1 & \textbf{21.7} \\
    SAMSum & Rouge-2 & Test & \citeauthor{narayan-etal-2021-planning}& 28.3 & \textbf{29.6} \\
    Gigaword & Rouge 2 & Test & \citeauthor{aghajanyan2020better} & \textbf{20.7} & \textbf{20.7} \\
    WebNLG (en) & Rouge-2 & Test & \citeauthor{bakshi2021structure} & 53.5 & \textbf{55.4}  \\ 
    E2E-NLG & Rouge-2 & Test & \citeauthor{xue2020mt5} & 45.8 & \textbf{46.5}\\
   CommonGen & Rouge-2 & Dev & \citeauthor{gehrmann2021gem}  & 32.5 & \textbf{37.4}\\
   Schema-Guided Dialog & Rouge-2 & Test & \citeauthor{gehrmann2021gem} & 36.8 & \textbf{44.1}\\
   \\
   
   GENIE - aNLG & Human (H) & Test & \citeauthor{genie} & 76.0 & \textbf{77.0}$^{(l)}$  \\
   GENIE - ARC-DA (w/o IR) & Human & Test & \citeauthor{khashabi2020unifiedqa} & \textbf{72.0} & \textbf{72.0}$^{(l)}$  \\
   GENIE - WMT19 & Human & Test & \citeauthor{genie} & \textbf{71.0} & 67.0$^{(l)}$\footnote{This task is German-to-English translation. Our submission is pretrained on only English C4 then finetuned on only the provided WMT19 data (no German pretraining, parallel data or backtranslation.)}  \\
   GENIE - XSUM & H-Overall & Test & \citeauthor{controlprefixes} & \textbf{51.0} & 50.0$^{(l)}$  \\
   GENIE - XSUM & H-Concise & Test & \citeauthor{controlprefixes} & \textbf{53.0} & \textbf{53.0}$^{(l)}$  \\  
   GENIE - XSUM & H-Fluency & Test & \citeauthor{controlprefixes} & 51.0 & \textbf{52.0}$^{(l)}$  \\
   GENIE - XSUM & H-No-Hallucination & Test & \citeauthor{controlprefixes} & 53.0 & \textbf{54.0}$^{(l)}$  \\
   GENIE - XSUM & H-Informativeness & Test & \citeauthor{controlprefixes} & \textbf{49.0} & \textbf{49.0}$^{(l)}$  \\
   \\
   
    SIQA & Accuracy & Test & \citeauthor{lourie2021unicorn} & 83.2 &  \textbf{83.3}$^{(l)}$  \\ 
    PIQA & Accuracy & Test & \citeauthor{lourie2021unicorn}  & 90.1 & \textbf{90.7}$^{(l)}$\\ 
    CSQA & Accuracy & Dev & \citeauthor{lourie2021unicorn}  & 79.1 & \textbf{84.9} \\ 
    CSQA2 & Accuracy & Test & \citeauthor{lourie2021unicorn} & 69.6$^{(\sharp)}$ & \textbf{70.1}$^{(l)}$ \\
    QASC (w/o IR) & Accuracy & Dev & \citeauthor{khashabi2020unifiedqa}& 81.8 & \textbf{83.8} \\
    QASC (w IR) & Accuracy & Test & \citeauthor{khashabi2020unifiedqa} & 89.6 & \textbf{90.7}$^{(l)}$ \\ 
    TweetQA & BLEU-1 & Dev & \citeauthor{khashabi2022unifiedqa} & 77.5 & \textbf{78.4} \\
   
    QuAIL & Accuracy & Test & \citeauthor{khashabi2022unifiedqa} & 74.2 & \textbf{87.2}\\ 
    AdversarialQA (Bert) & F1 & Dev &  \citeauthor{khashabi2022unifiedqa} & 53.6 & \textbf{70.1}\\
    AdversarialQA (Roberta) & F1 & Dev & \citeauthor{khashabi2022unifiedqa} & 45.5 & \textbf{57.5} \\
    AdversarialQA (Bidaf) & F1 & Dev & \citeauthor{khashabi2020unifiedqa} & 71.5 & \textbf{77.5}\\
    MCScript & Accuracy & Test & \citeauthor{khashabi2022unifiedqa} & 95.1 & \textbf{97.3} \\
    MCScript 2.0 & Accuracy & Test & \citeauthor{khashabi2022unifiedqa} & 94.6 & \textbf{97.9} \\ 
       RACE & Accuracy & Test &    \citeauthor{shoeybi2019megatron}  &\textbf{90.9}$^{(e)}$  & \textbf{90.9} \\ 
        DREAM & Accuracy & Test & \citeauthor{wan2020multi} & \textbf{91.8} & \textbf{91.8 }\\
          OBQA & Accuracy & Test & \citeauthor{khashabi2020unifiedqa}  & \textbf{87.2} & \textbf{87.2}$^{(l)}$\\
         CosmosQA & Accuracy & Test & \citeauthor{lourie2021unicorn}  &\textbf{91.8} & 91.6$^{(l)}$  \\
         Winogrande XL & Accuracy & Test & \citeauthor{lourie2021unicorn} & \textbf{91.3} & 90.1$^{(l)}$\\
    \\
        DocNLI & Accuracy & Test &\citeauthor{qin2022nlp} & 76.9 & \textbf{88.2} \\
    AdversarialNLI (r3) & Accuracy & Test & \citeauthor{wang2020infobert}  & 47.7 & \textbf{53.5}  \\
    VitaminC & Accuracy & Test & \citeauthor{schuster2021get}  & 90.8 & \textbf{91.1} \\ 
    Hellaswag & Accuracy & Test & \citeauthor{lourie2021unicorn} &  93.9 & \textbf{94.1}$^{(l)}$ \\
     QQP & F1 & Dev & \citeauthor{raffel2019exploring} & 90.1 & \textbf{90.6}\\
     QNLI & Accuracy & Dev & \citeauthor{raffel2019exploring} & 96.1 & \textbf{96.5} \\ 
     CoLA & Matthews & Dev & \citeauthor{raffel2019exploring} & 68.6 & \textbf{71.5} \\
     STSB & Spearman & Dev & \citeauthor{raffel2019exploring} & 92.1 & \textbf{92.3} \\ 
       AbductiveNLI & Accuracy & Test & \citeauthor{he2020deberta} &  \textbf{89.8}$^{(\sharp)}$ & 87.5$^{(l)}$ \\
       MultiNLI & Accuracy & Dev & \citeauthor{raffel2019exploring} & \textbf{92.1} & 91.9\\ 
      
       \\
     IMDB & Accuracy & Test & \citeauthor{yang2019xlnet} &96.2 &  \textbf{97.3} \\ 
    AgNews & Error & Test & \citeauthor{yang2019xlnet} &4.45  &  \textbf{4.42} \\ 
    Civil Comments & F1 & Dev & \citeauthor{tay2021pre} & 87.8 & \textbf{87.9}\\
    Wikipedia Toxicity & F1 & Dev & \citeauthor{tay2021pre}  & 96.5 & \textbf{97.0} \\
    SST-2 & Acc & Dev & \citeauthor{raffel2019exploring} & \textbf{97.3} & 97.0 \\
    \\
    Scrolls Challenge & Aggregate & Test & \citeauthor{shaham2022scrolls} &  29.2 & \textbf{37.9}$^{(l)}$ \\
    SumScr & Rouge (Avg) & Test & \citeauthor{shaham2022scrolls} & 16.3 & \textbf{20.0}$^{(l)}$   \\
    QMSum & Rouge (Avg) & Test & \citeauthor{shaham2022scrolls} & 19.9 & \textbf{20.0}$^{(l)}$ \\ 
    QASPER & F1 & Test & \citeauthor{shaham2022scrolls}  & 26.6& \textbf{37.6}$^{(l)}$ \\  
    NarrativeQA & F1 & Test & \citeauthor{shaham2022scrolls}  & 18.5& \textbf{24.2}$^{(l)}$\\
    QUALITY & EM & Test & \citeauthor{shaham2022scrolls}   & 26.0 & \textbf{45.8}$^{(l)}$  \\
    ContractNLI & EM & Test & \citeauthor{shaham2022scrolls}  & 77.4 & \textbf{88.7}$^{(l)}$ \\
     GovRep & Rouge (Avg) & Test & \citeauthor{shaham2022scrolls}  & \textbf{37.2} & 36.2$^{(l)}$\\ 
    \\

    WikiTQ & Accuracy & Test & \citeauthor{xie2022unifiedskg}  & 49.3 & \textbf{54.6} \\
    CompWebQ & Accuracy & Test &  \citeauthor{xie2022unifiedskg}  & 73.3 & \textbf{75.9} \\
    FetaQA & BLEU-4 & Test & \citeauthor{xie2022unifiedskg}  & 33.4 & \textbf{35.8} \\ 
    HybridQA & Accuracy & Dev & \citeauthor{eisenschlos2021mate}  & 60.8 & \textbf{61.0}\\
    WikiSQL & Accuracy & Test & \citeauthor{xie2022unifiedskg}  & 86.0 & \textbf{87.3} \\
    TabFat & Accuracy & Test & \citeauthor{xie2022unifiedskg}  & 83.4 & \textbf{87.1} \\
    Feverous & Accuracy & Dev & \citeauthor{xie2022unifiedskg}  & 82.4 & \textbf{85.6} \\
    SQA & Sent.Acc & Test & \citeauthor{xie2022unifiedskg}  & 62.4 & \textbf{70.5}\\
    MTOP & Match & Test & \citeauthor{xie2022unifiedskg}  & 86.8 & \textbf{87.5} \\
    DART & BLEU-4 & Test & \citeauthor{aghajanyan2021htlm}  & 47.2 &  \textbf{50.4} \\
    \\
    DSI-NQ & HITS@10 & Dev &\citeauthor{tay2022transformer}  & 70.3 & \textbf{73.8} \\ 
    \bottomrule
\end{longtable}
\end{center}
\normalsize

\subsubsection{Results on Supervised Finetuning} Our experimental results show that UL2 achieves state-of-the-art performance on around 50+ NLP tasks and setups. For many, the margins are quite wide and for those that UL2 doesn't achieve SOTA, the performance of UL2 is generally quite competitive. It is worth to note that the extent of difficulty of obtaining sota on each benchmark has vastly different difficulties. For some, the sota model is a 32B dense equivalent \citep{zoph2022designing}. For some others, it's a base model. It is worth also noting that many benchmarks have a strong relatively large model, e.g., 3B or 11B T5, UnifiedQA \citep{khashabi2020unifiedqa} or Unicorn \citep{lourie2021unicorn} as the existing SOTA model so outperforming these models is also not exactly the easiest thing to do. Overall, we urge the readers to judge the value of these sota results for themselves. Finally, we note that UL2 20B does pretty well on human evaluation on GENIE tasks, outperforming sota on several metrics. This ascertains that the generation quality of UL2 is reasonably solid.

\subsubsection{Tradeoffs between Finetuning and Prompt-based Zero-shot Learning (SuperGLUE)}
In this section, we explore finetuning and in-context learning trade-offs on the SuperGLUE benchmark. 
We conduct experiments on SuperGLUE with UL20B. While UL20B does not achieve SOTA on this benchmark, we note that UL20B at least remains competitive and outperforms T5-11B. This section reassures that UL2 indeed scales and matches/slightly outperforms T5-11B on SuperGLUE (while strongly outperforming T5-XXL on many other in-context tasks). UL20B still lacks behind the SOTA model ST-MoE-32B given two main reasons. Firstly, ST-MoE-32B has 200B+ parameters and is costs equivalent to a 32B dense model. Secondly, ST-MoE-32B is trained solely on span corruption using an encoder-decoder architecture which is known to be very advantageous on NLU finetuning.

\begin{table}[h!]
    \centering
    \small
        \caption{Results on SuperGLUE dev set. We compare with T5-11B \citep{raffel2019exploring}, ST-MoE-32B \citep{zoph2022designing} and PaLM-8B, PaLM-62B and PaLM-540B \citep{chowdhery2022palm}. Scores reported are the peak validation scores per task.}
    \label{tab:finetune-superglue}
    \begin{tabular}{lcccccccccc}
    \toprule
  Model      & BoolQ & CB & CoPA & MultiRC & Record & RTE & WiC & WSC & Avg   \\
  \midrule
  PaLM 62B & 90.6 & 96.4/95.7  & 98.0 & 87.7/61.9 & 93.0/92.4 &  89.5 & 75.9 & 96.2 & 89.2  \\
      PaLM 540B & 92.2 & 100/100 & 100 & 90.1/69.2 & 94.0/94.6 & 95.7 & 78.8 & 100  & 92.6 \\
     ST-MoE 32B$_{269B}$  & \textbf{93.1} & \textbf{100}/\textbf{100} & \textbf{100} & \textbf{90.4}/\textbf{69.9} & \textbf{95.0}/\textbf{95.6} & \textbf{95.7} & \textbf{81.0} & \textbf{100} & \textbf{93.2} \\
     \midrule 
     PaLM 8B  & 87.6 & 96.4/92.1 & 86.0 & 81.6/64.0 & 89.7/89.3 & 84.5 & 73.4 & 88.5 & 83.4\\
    T5 11B      & 90.8 & 94.9/96.4 & 98.0 & 87.4/66.1 & 93.8/93.2 & 93.9 & 77.3 & 96.2 & 89.9 \\
     UL2 20B & 90.8 & 98.7/98.2 & 99.0 & 88.4/64.8 & 93.7/93.2 & 92.1 & 77.3 & 98.1 &  90.7  \\
         \bottomrule
    \end{tabular}
\end{table}
  
\begin{table*}[h!]
    \centering
       \caption{Results on zero-shot learning on SuperGLUE dataset. We compare with GPT-3, GLaM and PaLM \citep{chowdhery2022palm}. We also include models that are relatively compute-matched with UL20B such as T5-XXL with LM adaptation \citep{lester2021power}, GPT-3 13B and GLaM-8B dense. Notably, UL20B outperforms GPT-3 175B and all other models in a similar compute class on average score.}
    \label{tab:my_label}
    \small
    \begin{tabular}{lccccccccc}
    \toprule
    Model & BoolQ & CB & RTE & ReCORD & WSC & WiC & COPA & MultiRC & Avg \\ 
    \midrule
    ST-MoE-32B (269B) & 40.8 & 41.1 & 52.7 & 50.0 & 57.5 & 50.0 &56.0 & 30.3 & 47.6 \\
     GPT-3 175B    &  60.5 &	46.4&	63.5&	90.2&	65.4&	0.0 &	91.0 &	72.9 &	61.2\\
       GLaM-MoE 1.2T   & 83.0 &	33.9 &	68.8 &		90.3 &		84.9&		50.5&		90.0&		45.1&		68.3\\
       PaLM 540B & 88.0	& 51.8 &	72.9 &	92.9 &	89.1 &	59.1 &	93.0 &	83.5 &	78.8\\
       \midrule
       T5-XXL & 44.3 & 	37.5 &	48.8 &	85.8 &	59.3 &	50.9 &	70.0 &	23.0 &	52.5\\
       GPT-3 13B & 66.2	& 19.6 & 	62.8 & 	89.0	& 64.4& 	0.0 & 	84.0 & 	71.4& 	57.2 \\
        GLaM-Dense 8B & 73.6 & 	33.9 &	44.0 &	89.2 &	80.7 &	44.0 &	86.0 & 	39.0 &	61.3\\
        GLaM-MoE 64E & 72.2	& 40.7 & 	60.3 & 	88.9 &	81.8 & 	49.5 & 	86.0	& 52.4 &	66.5\\
        PaLM-Dense 8B & 68.3 &	41.1&	54.2&	87.8&	78.9&	47.0&	86.0&	47.5&	63.9 \\
        \midrule
        UL2 20B (\textit{single ckpt}) & 63.1 & 	41.1 &	60.7 &	88.1 &	79.9 &	49.8&	85.0 &	36.2&	63.0 \\
       UL2 20B (\textit{best}) & 63.1 &	50.0 &	60.7	& 88.1 &	80.6 &	55.2 &	88.0	& 36.2 &	65.2\\
      
       \bottomrule
    \end{tabular}
 
\end{table*}

\subsubsection{Generative Few-shot: XSUM Summarization}
Finally, we conduct additional few-shot in-context one-shot learning using the XSum dataset We compare our model with the baseline T5-XXL, T5-XXL with LM Adaptation \citep{lester2021power}, LaMDA 137B \citep{thoppilan2022lamda}, and PaLM (8B, 62B, 540B) \citep{chowdhery2022palm}. We run T5-XXL ourselves in the same experimental setup but report results from \citep{chowdhery2022palm} for the other models.

\begin{table}[H]
    \centering
    \small
      \caption{Results on One-Shot Summarization on XSUM.}
    \label{tab:1shot_xsum}
    \begin{tabular}{lccc}
    \toprule 
    Model     & Rouge-1 & Rouge-2 & Rouge-L  \\
    \midrule
    LaMDA 137B & - & 5.4 & -\\
    PaLM 62B & - & 11.2 & -\\
    PaLM 540B & - & \textbf{12.2} & -\\
    \midrule
    PaLM 8B & - & 7.9 & - \\
        T5 XXL 11B &  0.6 & 0.1 & 0.6 \\
        T5 XXL 11B + LM & 13.3 & 2.3 & 10.7 \\
        UL2 20B & \textbf{25.5} & \textbf{8.6} & \textbf{19.8}\\
        \bottomrule
    \end{tabular}
\end{table}
Table \ref{tab:1shot_xsum} reports results on 1-shot summarization. Our results show that the performance of UL2 20B is about 3x the performance of LM adapted T5 XXL model. Moreover, UL2 20B outperform LaMDA 137B and has better performance compared to PaLM 8B which is approximately compute-matched with UL2. The best result, however, is still the larger 540B and 62B PaLM models.

\subsubsection{UL2 for chain-of-thought prompting}
It has recently been shown that language models at scale can perform multi-step reasoning tasks such as math word problems or commonsense reasoning via \textit{chain-of-thought prompting}, which prompts the model to generate a step-by-step reasoning path before giving the final answer \citep{wei2022chain}.
Notably, chain-of-thought (CoT) prompting does not require any additional fine-tuning of the model.

A crucial consideration of CoT prompting is that it is an emergent ability of scale \citep{wei2022emergent}---it requires a sufficiently large language model to improve performance, and actually hurts performance for small language models. 
Hence, the successful use cases of chain-of-thought prompting use either LaMDA 137B \citep{thoppilan2022lamda}, PaLM 540B \citep{chowdhery2022palm}, or OpenAI models \citep{brown2020language,ouyang2022training}.
These models, however, are compute intensive and not available as public checkpoints.

Here we demonstrate that UL2 20B is the first publicly available pre-trained model (without any fine-tuning) to successfully leverage CoT prompting to solve multi-step arithmetic and commonsense tasks.
We use the same benchmark tasks and prompts from \citet{wei2022chain}.
In Table \ref{tab:cot-arithmetic} below, we see that on five arithmetic reasoning datasets, CoT prompting outperforms standard prompting (directly outputting the answer without a chain of thought) for UL2 20B.
Similar to \citet{wei2022chain}, we also show that CoT prompting can be augmented by using an external calculator (``calc.'') to perform arithmetic computational only ($+,-,\times,/$) to further improve performance by a large margin. In addition, we add self-consistency \citep{self_consistency} (denoted as ``SC'') on top of CoT prompting and observed significant gains consistently across all benchmarks, with an average improvement of 22.5\% compared to standard prompting.

\begin{table}[H]
    \centering
    \small
      \caption{Chain-of-thought prompting and self-consistency (SC) results on five arithmetic reasoning benchmarks.
      GSM8K: \citep{cobbe2021training}.
      SVAMP: \citep{patel-etal-2021-nlp}.
      ASDiv: \citep{miao-etal-2020-diverse}.
      AQuA: \citep{ling-etal-2017-program}.
      MAWPS: \citep{koncel-kedziorski-etal-2016-mawps}.
      }
    \label{tab:cot-arithmetic}
    \begin{tabular}{l ccccc c}
    \toprule 
    Model     & GSM8K & SVAMP & ASDiv & AQuA & MAWPS & \underline{Average} \\
    \midrule 
   UL2 20B: standard prompting & 4.1 & 10.1 & 16.0 & 20.5 & 16.6 & 13.5 \\
    UL2 20B: CoT prompting & 4.4 & 12.5 & 16.9 & 23.6 & 19.1 & 15.3 \\
    UL2 20B: CoT prompting + calc. & 6.9 & 28.3 & 34.3 & 23.6 & 42.7 & 27.2 \\
    UL2 20B: CoT prompting + calc. + SC & \textbf{10.2} & \textbf{41.4} & \textbf{43.5} & \textbf{26.9} & \textbf{57.9} & \textbf{36.0}\\
    \bottomrule
    \end{tabular}
\end{table}

In addition to arithmetic reasoning, Table \ref{tab:cot-commonsense} shows the performance of CoT prompting using UL2 20B compared to standard prompting on five commonsense reasoning benchmarks.
CoT prompting plus self-consistency outperforms standard prompting in four of the five benchmarks, with an average improvement of 14.4\%. 

\begin{table}[H]
    \centering
    \small
      \caption{Chain-of-thought prompting and self-consistency (SC) results on five commonsense reasoning benchmarks.
      CSQA: \citep{talmor-etal-2019-commonsenseqa}.
      StrategyQA: \citep{geva-etal-2021-aristotle}.
      Date Understanding and Sports Understanding: \citep{srivastava2022beyond}.
      ARC-easy/challenge: \citep{ARC}.
      }
    \label{tab:cot-commonsense}
    \begin{tabular}{l ccccccc }
    \toprule 
    Model     & CSQA & StrategyQA & Date & Sports & ARC-e & ARC-c & \underline{Average} \\
    \midrule 
  UL2 20B: standard prompting & 34.2 & \textbf{59.0} & 13.5 & 57.9 & 32.2  & 29.8 & 37.8  \\
    UL2 20B: CoT prompting & 51.4 & 53.3 & 14.0 & 65.3 & 61.6 & 42.9 & 48.1 \\
    UL2 20B: CoT prompting + SC & \textbf{55.7} & 54.9  & \textbf{16.3} &\textbf{66.8} & \textbf{69.8}  & \textbf{49.5} & \textbf{52.2}\\
    \bottomrule
    \end{tabular}
\end{table}

Overall, we have shown that whereas prior CoT work have required large pre-trained models such as PaLM 540B, UL2 20B is a relatively smaller model that can also perform multi-step reasoning. 
We hypothesize that the mixture of denoisers may contribute to the ability of UL2 to leverage CoT prompting at 20B parameters, although we leave further investigation of what unlocks emergent chain-of-thought reasoning to future work.

\subsubsection{Massively Multitask Language Understanding}

Massive Multitask Language Understanding (MMLU)~\citep{hendrycks2021_mmlu} is a collection of 57 tasks covering a wide range of topics (humanities, social sciences, hard sciences, etc.). Strong performance on MMLU requires extensive world knowledge as well as problem solving skills.

For MMLU, we compare with T5 model variants including the language model adapted variant~\cite{lester2021power} and T0~\citep{sanh2019distilbert}. For the latter, we use ``T0 Strawberry'' and ``T0 Vanilla" as these are the models recommended for research purposes. We report 0-shot performance. T0 models are specifically finetuned for 0-shot and hence we believe this is a conservative setting to test the efficacy of UL2. Table~\ref{table:mmlu} shows that the LM-adapted T5-XXL model barely gives above-random performance (25\%). UL2 outperforms both T0 and T5 models.

\begin{table}[H]
\begin{center}
\caption{MMLU 0-shot performance (accuracy).}
\label{table:mmlu}
\begin{tabular}{lc}
\toprule
                     & \multicolumn{1}{l}{MMLU} \\
                     \midrule
T5-XXL + LM  & 27.5                            \\
T0 Strawberry        & 36.9                            \\
T0 Vanilla           & 34.5                            \\
UL2 20B              & \textbf{39.2}                            \\
    \bottomrule
\end{tabular}
\end{center}
\end{table}

\subsection{Instruction Tuned UL2 20B with FLAN}
Inspired\footnote{Inspiration is really a stretch, this was an obvious thing to do.} by \citet{chung2022scaling}, we apply Flan instruction tuning on the UL2 20B checkpoint. We pretty much use the same settings and Flan mixture as the Flan2 paper \citep{chung2022scaling}. Because the flan mixtures do not have mode switching prompts, we opt to train UL2 for another 100K steps without mode tokens to adapt it. We increased the length to $1024/1024$ this time to accommodate a larger context length. Flan training was done at $2048/512$ length. We find the `mode switching purification' of the original UL2 checkpoint to be useful, although the more optimal way would to be to add mode tokens to the FLAN tasks. Since we were lazy to do that, we simply opt to continue training UL2 again for more steps. We release this Flan-UL2 20B checkpoint at the same url as the original UL2 checkpoints. 

\subsubsection{Few-shot MMLU and Big-Bench Results after Flan training of UL2}
\begin{table}[H]
    \centering
    \begin{tabular}{l|cc}
    \toprule
         &  BBH & MMLU \\
             \midrule
        OPT 30B & 28.0 & 23.5/25.9 \\
        OPT 175B & 30.2 & 27.3/34.2 \\
        T5 11B & 29.5 & -/25.9 \\
        OpenAI davinci & 33.6 & -/32.3 \\
        \midrule
        OPT IML-Max 30B & 30.0 & 46.3/43.2 \\ 
        OPT IML-Max 175B & 35.7 & 49.1/47.1 \\
        T0pp 11B & 13.0 & 46.7/33.7 \\
        FLAN T5 XXL & 45.3 & 54.5/53.7 \\
        FLAN-PaLM 62B & 47.5 & -/59.6 \\ 
        FLAN-PaLM 540B & 57.9 & -/73.5 \\
        \midrule
        FLAN-UL2 20B (Best ckpt for both tasks$^\dagger$) & 45.3 & 55.6/56.2 \\
        FLAN-UL2 20B (Individual task best) &46.0  & 55.1/58.1\\
        \bottomrule
    \end{tabular}
    \caption{Results on MMLU and BBH using FLAN-UL2. $\dagger$ denotes checkpoint that we release.}
    \label{tab:flan_results}
\end{table}

Table \ref{tab:flan_results} reports the results on MMLU and BBH \citep{suzgun2022challenging}. Generally, the performance of FLAN-UL2 20B is pretty competitive and outperforms Flan-T5 XXL by $+1.8\%$ on the test set and $+4.7\%$ on MMLU dev. The Big-Bench hard score remains competitive with the best checkpoint marginally outperforming Flan-T5 XXL. Notably, the best dev scores of FLAN-UL2 is almost reaching the performance of Flan-PaLM 62B on both MMLU and BBH, suggesting that the results are pretty solid.

\subsubsection{Comparisons on using Chain-of-thought vs Direct Prompting}
We compare Flan models on direct and chain-of-thought setup. We fine-tune Flan-UL2 using the exact identical protocol as T5-XXL and pick the best score based on the strongest average\footnote{This results slightly differ from the earlier setup because we release the best checkpoints only on direct prompting. CoT did worse, so we prioritized direct prompting.} across all four setups (MMLU/BBH with direct and CoT). We find that Flan-UL2 outperforms Flan-T5-XXL on all four tasks. Notably, there are larger gains on CoT tasks, e.g., especially MMLU-CoT where the gain is a relative +7.4\%. In general, CoT variants for these tasks still perform worse than direct which can also be observed in the PaLM 62B model. This also seems to be true for Flan-PaLM 62B. Overall, Flan-UL2 comes close to FLAN-PaLM 62B (49.1 vs 49.9) on average across all setups. However, it is still strongly outperformed by Flan-PaLM 540B. 
\begin{table}[H]
    \centering
    \begin{tabular}{l|ccccc}
    \toprule
         & MMLU & BBH & MMLU-CoT & BBH-CoT & Avg \\
        \midrule
           FLAN-PaLM 62B & 59.6 & 47.5 & 56.9&44.9 & 49.9  \\ 
           FLAN-PaLM 540B & 73.5 & 57.9 & 70.9 & 66.3 & 67.2\\
           \midrule
        FLAN-T5-XXL 11B & 55.1 & 45.3 & 48.6 & 41.4 & 47.6 \\
        FLAN-UL2 20B & 55.7 (+1.1\%) & 45.9 (+1.3\%) &  52.2 (+7.4\%) & 42.7 (+3.1\%) & 49.1 (+3.2\%)\\ 
        \bottomrule
    \end{tabular}
    \caption{Comparisons (dev scores) of Flan models on CoT vs Direct.}
    \label{tab:my_label}
\end{table}
We also tried some self-consistency \citep{self_consistency} experiments in combination with CoT. From brief experiments, this raised CoT score from 53.9 to 57.1 (when the corresponding direct score was 55.4). This shows that at 20B scale, CoT + self consistency can outperform direct prompting. We did not experiment further since this increases the search space to a point where it was more time consuming than we would have liked (or enjoyed). We leave any future experiments as an exercise for the reader. 
\section{Conclusion} 
We proposed a new paradigm for training universally effective models. UL2 is characterized by two key ideas. Firstly, we propose a new Mixture of Denoisers (MoD) pretraining that frames multiple pretraining tasks as span corruption, diversifies and then mixes them. Secondly, we introduce mode switching, a way of associating downstream task behaviour to upstream pretraining. Extensive ablative experiments show that UL2 consistently outperforms GPT-like and T5 models on a wide range of supervised and few-shot tasks, outperforming T5 on 9 out of 9 tasks and by a normalized overall gain of +76.1\%. Finally, we scale UL2 up to 20B parameters and conduct experiments on a diverse suite of 50 to 60 NLP tasks and setups. UL2 achieves sota performance on 50 of them. Pretrained checkpoints of UL2 and Flan-UL2 20B are released at \url{https://github.com/google-research/google-research/tree/master/ul2}.

\newpage 
\section{Acknowledgements}
The authors would like to specially thank (in alphabetical order): Alexey Gritsenko, Andrew M.  Dai, Jacob Devlin, Jai Gupta, Liam Fedus, Orhan Firat for discussions and feedback that have helped to improve the paper. We also thank Sebastian Gerhmann for discussions and clarifications on GEM metrics, Nan Du for clarifications about GLaM's in-context learning setup and Dave Uthus for his work on getting the scrolls tasks in seqio format. We thank Slav Petrov and Quoc Le for general advice about UL2. We also thank the T5, Jax and Flax teams for building such wonderful infrastructure and enabling this research. Finally, we also thank Tianbao Xie from University of Hong Kong for helping us with UnifiedSKG's code and dataset. 

\section{Author Contributions}
This section lists the author contributions of each author. 
\begin{itemize}
\item Yi Tay proposed the idea, conceived the project, lead this effort and drove the implementation, core ablation experiments. Yi ran initial ablations, proof-of-concepts and pretrained the 20B model. Yi was responsible for running most of the finetuning and in-context learning experiments for the 20B model. 

\item Mostafa Dehghani served as a co-lead of this effort and ran a good portion of initial experiments and ablations, especially on SuperGLUE. Mostafa was quite involved in early brainstorming of this effort and project. Mostafa helped out on the open sourcing process and procedures of UL2.

\item Vinh Q. Tran participated substantially in early project discussions and brainstorming and contributed to the the inception of UL2. Vinh implemented and trained UL2 on several tasks/baselines (e.g., SamSum, GENIE human evaluations, CommonsenseQA) for the UL2 20B runs. 

\item Xavier Garcia helped optimize our the UL2 pipeline in seqio and provided many great suggestions about optimizing UL2. Xavier also ran experiments on UL2 in machine translation. 
\item Jason Wei ran Chain-of-thought experiments on reasoning benchmarks using the UL2 model.
\item Xuezhi Wang ran self-consistency experiments on reasoning benchmarks using the UL2 model.
\item Hyung Won ran experiments for the MMLU dataset and wrote the section for it. 
\item Siamak extensively helped with UL2 experiments, infrastructure and continiously improving the UL2 algorithm.
\item Dara Bahri helped port UnifiedSKG for UL20B sota experiments. 
\item Tal Schuster helped out with UL20B evaluations on the Scrolls leaderboard. Tal also helped with evaluating UL20B on VitaminC and Programming Puzzles datasets.

\item Huaixiu Steven Zheng brainstormed and discussed the idea with Yi and helped to write the paper and provide feedback. 
\item Denny Zhou suggested running chain of thought and reasoning experiments with UL2, helped advise the chain-of-thought section. 
\item Neil and Donald served as technical advisors and sponsors to the project and helped brainstorm, provide feedback and writing of the paper. 
\end{itemize}

\bibliography{ref}
\bibliographystyle{iclr2021_conference}

\newpage
\section{Appendix}

\subsection{Model Release}
As part of this work, we release the weights of the 20B checkpoint. The weights of the model can be found in in this GCP bucket ({\url{gs://scenic-bucket/ul2}}). These checkpoints were trained with T5X \citep{https://doi.org/10.48550/arxiv.2203.17189} found at  \url{https://github.com/google-research/t5x} and are implemented in JAX/Flax. Because the fine-tuning results are generally not from a single checkpoint due to our continuous finetuning setup, we release three different checkpoints (1.87M, 2.05M, 2.65M) which we found to be consistently pretty good.

A slight \textit{disclaimer} is that we finetuned and trained this model on TPUv4 chips on our internal systems. Even so, finetuning would also sometimes result in \textit{nans} which may require some care and manual tuning to get resolved. Therefore, if checkpoints were to be ported to another system, we could not  guarantee that these checkpoints would work as well. We are overall optimistic but we do not guarantee its stability with external hardware and accelerators such as GPUs.

For this particular checkpoint, note that the mode tags we used are [NLG] (X-denoiser), [NLU] (R-denoiser) and [S2S] (S-denoiser). So add that at the start of the inputs of your examples. 
\subsection{Implementation Details and UL2 code}
This section aims to give more insight to how UL2 pretraining is implemented. Our implementation is actually pretty simple. It is simply a mixture of different pretraining objectives that is implemented in seqio\footnote{\url{https://github.com/google/seqio}}. Most of our experiments were run with simply mixing different seqio tasks with seqio's Mixture Registry. However, one could also implement a generalized UL2 objective with the following function which could be cleaner.
\begin{lstlisting}
def ul2_objective(dataset: tf.data.Dataset,
                    sequence_length: seqio.preprocessors.SequenceLengthType,
                    output_features: seqio.preprocessors.OutputFeaturesType,
                    use_prefix_lm_task: bool = False,
                    rates: Optional[Sequence[float]] = None,
                    mean_noise_span_lengths: Sequence[float] = (3.0,),
                    noise_densities: Sequence[float] = (0.15,),
                    shard_ds: bool = True,
                    optional_task_prefixes: Optional[Sequence[str]] = None,
                    input_feature_key: str = "inputs",
                    merge_examples_to_reduce_padding: bool = True,
                    reserved_for_packing: bool = None,
                    seed: int = 7) -> tf.data.Dataset:
  """UL2-like pre-training objectives.

  This preprocessor amounts to calling the `span_corruption` function several
  times with different values of `noise_density` and `mean_noise_span_length`.
  We either shard or copy the dataset, then apply each function to each shard.
  Add S-denoising (prefixLM) using use_prefix_lm_task. 

  Args:
    dataset: A tf.data.Dataset with dictionaries containing the key
      `input_feature_key`.
    sequence_length: dict mapping of feature key to int length for that feature.
    output_features: mapping of keys to features.
    use_prefix_lm_task: <bool> If True, include PrefixLM in the task mix.
    rates: <Optional<List<float>> List of rates per task. If None, tasks are
           sampled uniformly.
    mean_noise_span_lengths: List of mean number of tokens per masked span per
      example.
    noise_densities: List of what fraction of the tokens to mask.
    shard_ds: <bool> If True, shard dataset per objective.
    optional_task_prefixes: <Optional<list<str>> Strings to prepend for each
                            corruption scheme. NOTE: If including prefixLM task,
                            it must be the last prefix.
    input_feature_key: which feature to use from the dataset as the input text
      tokens.
    merge_examples_to_reduce_padding: if True, combines multiple input examples
      to reduce padding.
    reserved_for_packing: if specified, reduces the desired inputs length by the
      specified amount to enable multiple examples to be packed together
      downstream.
    seed: tf.int64 for controlling the random choice of spans.

  Returns:
    a dataset
  """

  if optional_task_prefixes:  # Ensure each task has a prefix.
    num_tasks = len(noise_densities) + int(use_prefix_lm_task)
    valid_number_of_prefixes = num_tasks == len(optional_task_prefixes)
    if not valid_number_of_prefixes:
      raise ValueError("Number of task prefixes must match number of tasks.")
  inputs_length = sequence_length[input_feature_key]
  input_lengths, targets_lengths = [], []
  sequence_lengths = {x: y for x, y in sequence_length.items()}
  if reserved_for_packing:
    inputs_length -= reserved_for_packing
    for x, y in sequence_length.items():
      sequence_lengths[x] = y - reserved_for_packing
  hyperparams = list(zip(mean_noise_span_lengths, noise_densities))
  for mean_noise_span_length, noise_density in hyperparams:
    input_length, targets_length = t5.data.preprocessors.random_spans_helper(
        extra_tokens_per_span_inputs=1,
        extra_tokens_per_span_targets=1,
        inputs_length=inputs_length,
        mean_noise_span_length=mean_noise_span_length,
        noise_density=noise_density)
    input_lengths.append(input_length)
    targets_lengths.append(targets_length)

    if sequence_length["targets"] < targets_length:
      upper_bound = max(targets_lengths)
      raise ValueError(
          f'Expected max targets length for span corruption ({upper_bound}) is '
          f'greater than configured targets length '
          f"({sequence_length['targets']})")

  ds = dataset
  ds = t5.data.preprocessors.select_random_chunk(
      ds,
      output_features=output_features,
      feature_key="targets",
      max_length=65536)
  if merge_examples_to_reduce_padding:
    ds = t5.data.preprocessors.reduce_concat_tokens(
        ds, feature_key="targets", batch_size=128)
  num_shards = len(input_lengths) + int(use_prefix_lm_task)
  if shard_ds:
    ds_shards = [ds.shard(num_shards, i) for i in range(num_shards)]
  else:
    ds_shards = [ds for _ in range(num_shards)]
  processed_ds = []
  hyperparams = zip(input_lengths, hyperparams, range(num_shards))
  for input_length, (noise_span_length, noise_density), i in hyperparams:
    ds = ds_shards[i]
    ds = t5.data.preprocessors.split_tokens(
        ds,
        feature_key="targets",
        min_tokens_per_segment=None,
        max_tokens_per_segment=input_length)
    ds = t5.data.preprocessors.denoise(
        ds,
        output_features,
        inputs_fn=t5.data.preprocessors.noise_span_to_unique_sentinel,
        targets_fn=t5.data.preprocessors.nonnoise_span_to_unique_sentinel,
        noise_density=noise_density,
        noise_mask_fn=functools.partial(
            t5.data.preprocessors.random_spans_noise_mask,
            mean_noise_span_length=noise_span_length),
        input_feature_key=input_feature_key)
    if optional_task_prefixes:
      ds = prepend_prompt(
          ds,
          output_features,
          prompt_mode=optional_task_prefixes[i],
          mode=optional_task_prefixes[i])
    processed_ds.append(ds)
  if use_prefix_lm_task:
    ds = ds_shards[-1]
    ds = t5.data.preprocessors.prefix_lm(ds, sequence_lengths, output_features)
    if optional_task_prefixes:
      ds = prepend_prompt(
          ds,
          output_features,
          prompt_mode=optional_task_prefixes[-1],
          mode=optional_task_prefixes[-1])
    processed_ds.append(ds)

  ds = tf.data.experimental.sample_from_datasets(processed_ds, rates, seed)
  return ds
\end{lstlisting}

\subsection{Details of Supervised Finetuning SOTA runs}
Most of our supervised finetuning runs were finetuned as single tasks. The only exception was that:
\begin{itemize}
    \item We finetuned GLUE as a single mixture with proportionate sampling. This has become standard and defacto setup \citep{raffel2019exploring,he2022hyperprompt,tay2020hypergrid,tay2021scale}.
    \item We finetuned SuperGLUE as a single mixture which is also a standard setup these days \citep{fedus2021switch,raffel2019exploring,chowdhery2022palm}.
    \item SIQA, PIQA, AbductiveNLI, Winogrande XL and CosmosQA were co-trained in a proportionate mixture similar to \citep{lourie2021unicorn} under the Rainbow benchmark.
    \item For CSQA, CSQA2. OBQA, and ARC-DA we co-trained with the rainbow mixture to obtain results on these three datasets. 
    \item All other tasks were single-task finetuned. 
\end{itemize}

\subsection{Details of Prompts for few-shot and zero-shot}
We report the optimal prompt for the zero-shot SuperGLUE experiments.
\begin{table}[H]
    \centering
    \begin{tabular}{c|c}
    \toprule
        Task &  Prompt \\
        \midrule
        BoolQ  & S2S \\ 
        CB & NLU \\ 
        RTE & S2S \\ 
        Record & S2S \\ 
        WiC & S2S \\
        WSC & S2S \\
        COPA & NLU \\
        MultiRC & S2S \\
        \bottomrule
    \end{tabular}
    \caption{Caption}
    \label{tab:my_label}
\end{table}

\newpage

\end{document}